\documentclass[sigconf]{acmart}

\AtBeginDocument{%
  }

\setcopyright{acmlicensed}
\copyrightyear{2026}
\acmYear{2026}
\acmPrice{}
\acmDOI{}
\acmISBN{}
\acmConference[KDD '26]{Proceedings of the 32nd ACM SIGKDD Conference on Knowledge Discovery and Data Mining V.1}{August 09--13, 2026}{Jeju Island, Republic of Korea}

\usepackage{balance}
\usepackage{threeparttable}
\usepackage{booktabs}
\usepackage{multirow}
\usepackage{enumitem}
\usepackage{xspace}
\usepackage[caption=false,font=footnotesize]{subfig}
\usepackage{makecell}

\newcommand{\model}[0]{FaST\xspace}
\newcommand{\mat}[1]{{\bf #1}}
\renewcommand{\vec}[1]{{\bf #1}}

\begin{document}


\title[\model: Efficient and Effective Long-Horizon Forecasting for Large-Scale Spatial-Temporal Graphs]{\model: Efficient and Effective Long-Horizon Forecasting for Large-Scale Spatial-Temporal Graphs via Mixture-of-Experts}


\author{Yiji Zhao}
\affiliation{
  \institution{Yunnan University}
  \department{School of Information Science and Engineering}
  \city{Kunming}
  \country{China}}
\email{yjzhao@ynu.edu.cn}

\author{Zihao Zhong}
\affiliation{
  \institution{Yunnan University}
  \department{School of Information Science and Engineering}
  \city{Kunming}
  \country{China}}
\email{zhongzihao@stu.ynu.edu.cn}

\author{Ao Wang}
\affiliation{
  \institution{Yunnan University}
  \department{School of Information Science and Engineering}
  \city{Kunming}
  \country{China}}
\email{wangao@stu.ynu.edu.cn}

\author{Haomin Wen}
\affiliation{
  \institution{Carnegie Mellon University}
  \city{Pittsburgh}
  \country{United States}}
\email{haominwe@andrew.cmu.edu}

\author{Ming Jin}
\affiliation{
  \institution{Griffith University}
  \city{Brisbane}
  \country{Australia}}
\email{mingjinedu@gmail.com}

\author{Yuxuan Liang}
\affiliation{
  \institution{The Hong Kong University of Science and Technology (Guangzhou)}
  \city{Guangzhou}
  \country{China}}
\email{yuxliang@outlook.com}

\author{Huaiyu Wan}
\affiliation{
  \institution{Beijing Jiaotong University}
  \city{Beijing}
  \country{China}}
\email{hywan@bjtu.edu.cn}

\author{Hao Wu}
\authornote{Hao Wu is the corresponding author.}
\affiliation{
  \institution{Yunnan University}
  \department{School of Information Science and Engineering}
  \city{Kunming}
  \country{China}}
\email{haowu@ynu.edu.cn}

\renewcommand{\shortauthors}{Yiji Zhao et al.}

\begin{abstract}
    Spatial-Temporal Graph (STG) forecasting on large-scale networks has garnered significant attention. However, existing models predominantly focus on short-horizon predictions and suffer from notorious computational costs and memory consumption when scaling to long-horizon predictions and large graphs. Targeting the above challenges, we present \textbf{\model}, an effective and efficient framework based on heterogeneity-aware Mixture-of-Experts (MoEs) for long-horizon and large-scale STG forecasting, which unlocks one-week-ahead (672 steps at a 15-minute granularity) prediction with thousands of nodes. \model is underpinned by two key innovations. First, an adaptive graph agent attention mechanism is proposed to alleviate the computational burden inherent in conventional graph convolution and self-attention modules when applied to large-scale graphs. Second, we propose a new parallel MoE module that replaces traditional feed-forward networks with Gated Linear Units (GLUs), enabling an efficient and scalable parallel structure. Extensive experiments on real-world datasets demonstrate that \model not only delivers superior long-horizon predictive accuracy but also achieves remarkable computational efficiency compared to state-of-the-art baselines. 
    Source code is publicly available at: \url{https://github.com/yijizhao/FaST}.
\end{abstract}



\maketitle

\section{Introduction}
Spatial-Temporal Graph (STG) serves as a powerful data schema for modeling complex dependencies in urban sensing systems, where nodes represent sensors (e.g., traffic detectors, energy meters) and edges capture spatial relationships~\cite{rahmani2023graph,JinLFSHZZ24}. Accurate long-horizon STG forecasting (e.g., days ahead) can shift urban management from reactive responses to proactive planning. For example, a seven-day electricity demand outlook allows renewable generation and storage to be scheduled ahead of peak loads. Such scenarios are typically coupled with two features: 1) a long forecasting horizon (e.g., 672 steps at 15-minute granularity covers one week) and 2) a large-scale graph (e.g., over thousands of nodes).

\begin{figure}[t]
  \centering
  \includegraphics[width=0.9\linewidth]{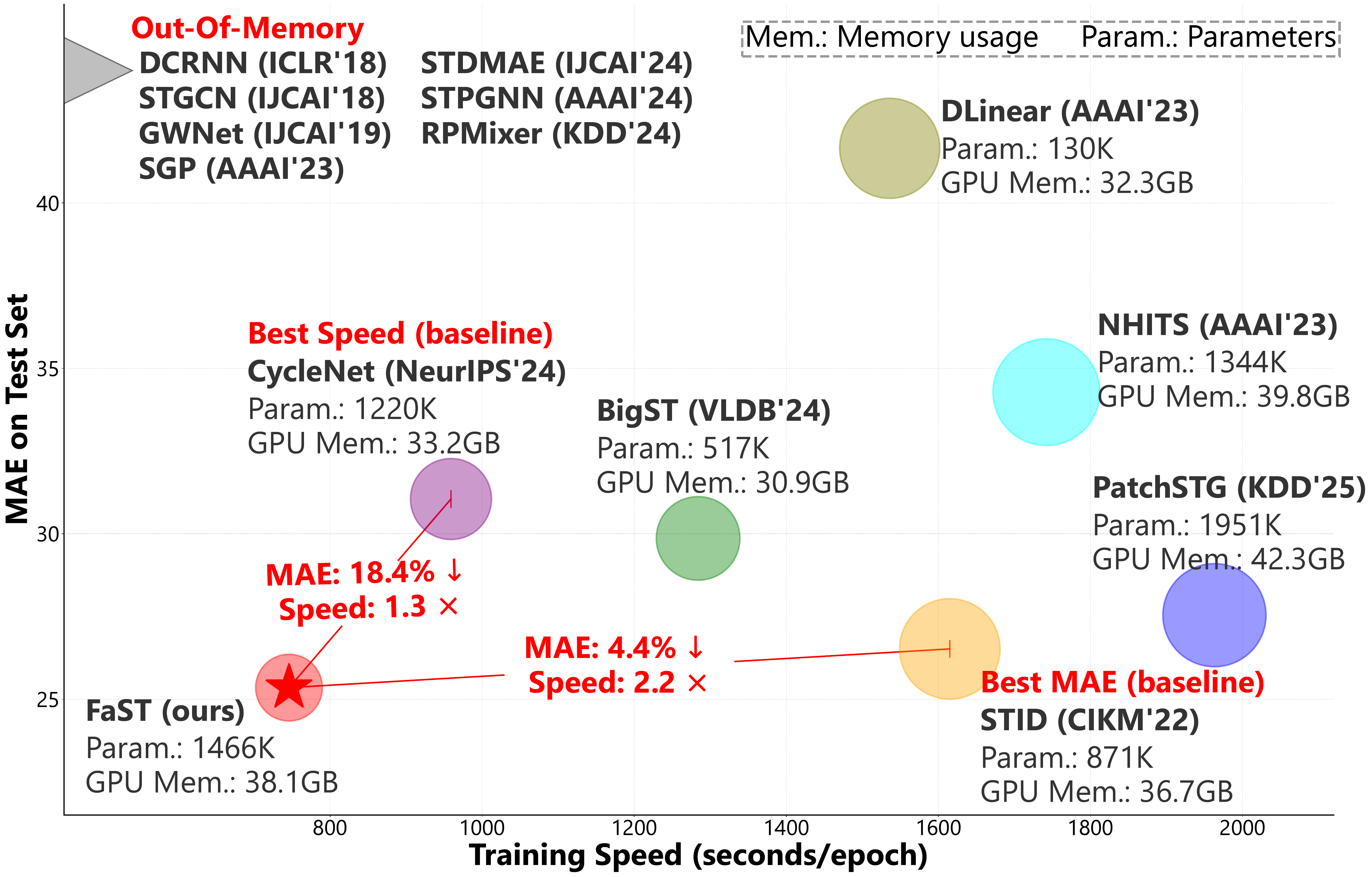}
  \caption{Efficiency--effectiveness comparison on large-scale STG benchmark (CA dataset; 8,600 nodes; 672-step prediction horizon). Smaller bubble means faster inference speed. The proposed \model achieves the best performance and speed (both in training and inference).}
  \Description{}
  \label{fig:intro}
\end{figure}

Extensive efforts have been made in STG forecasting, where most works focus on short-term forecasting (i.e., 12-step) in graphs with fewer than hundreds of nodes. Representative models build a Spatial-Temporal Graph Neural Networks (STGNNs), such as DCRNN~\cite{DCRNN} and STGCN \cite{STGCN}, coupling Graph Neural Networks (GNNs) with sequence models to capture both the spatial and temporal dependencies. Later improvements incorporate graph attention~\cite{ASTGCN,GMAN,ASTGNN}, temporal transformers~\cite{ASTGNN,Airformer}, and neural graph ODEs~\cite{STGODE} to boost short-term accuracy. However, those STGNNs suffer from quadratic complexity hidden in both spatial and temporal modules: GNNs and spatial/temporal attention incur $O(N^2)$ and $O(T^2)$ pairwise node/time-step interactions over $N$ nodes and sequence length $T$, respectively. When $N$ and $T$ scale jointly, resource demands increase exponentially. To give a concrete example, we report the training cost vs. performance on a large-scale STG benchmark \cite{LargeST} in Figure~\ref{fig:intro}, when confronted with 8,600 nodes and a 672-step prediction horizon (a practical setting in CA dataset), most representative STGNNs exhaust 48 GB of GPU memory, making them impractical to train and deploy in most GPU devices.

To elevate the STGNNs for large-scale and long-horizon forecasting, \textbf{\textit{a critical challenge is to achieve linear computational complexity while retaining the rich spatial and temporal semantics required for long-horizon forecasting.}} Recent studies have explored two complementary directions to reduce pairwise node/time-step interactions, though promising, they still face major limitations in both performance and efficiency.

\textbf{Spatially}, existing efficient methods can be broadly categorized into \textit{structure-aware} and \textit{structure-free} approaches. Structure-aware methods leverage the explicit graph topology to reduce spatial interactions through techniques such as sparse aggregation (e.g., SGP~\cite{SGP}), neighbor sampling (e.g., SAGDFN~\cite{SAGDFN}), and graph partitioning (e.g., PatchSTG~\cite{PatchSTG}). However, these approaches often discard long-range dependencies and heavily rely on an accurate graph structure (which may not always be available in real-world scenarios). To circumvent this, structure-free methods avoid using graph structure and instead rely on node embeddings and feature-mixing mechanisms. For example, BigST~\cite{BigST} adopts linear attention to bypass pairwise node computation; RPMixer~\cite{RPMixer} employs random projection and MLPs to facilitate efficient spatial fusion; and STID~\cite{STID} eliminates spatial interactions by encoding spatial identity with positional embeddings. While these techniques reduce the complexity from $O(N^2)$ to $O(N\cdot \phi)$, where $\phi\textless N$ is a tunable factor depending on the implementation and hyperparameters (e.g., sampled neighbors, feature dimension, or partition granularity), they often oversimplify the interaction operation, which can dilute meaningful relationships, introduce irrelevant noise, and therefore lead to significant loss of spatial semantics.

\textbf{Temporally}, two main strategies have emerged: \textit{sparse/linear attention mechanisms} and \textit{temporal pattern compression methods}. The former, adopted in models like Airformer~\cite{Airformer}, SAGDFN~\cite{SAGDFN}, and BigST~\cite{BigST}, aims to alleviate the quadratic cost of temporal attention by approximating it with lower-complexity variants. The latter strategy, employed in methods such as CycleNet~\cite{CycleNet} and STID~\cite{STID}, compresses long historical sequences into a compact low-dimensional embedding per node, effectively removing explicit pairwise interactions across time steps. For example, STID directly compresses the $T$-step history into a dense embedding via a simple linear projection, while CycleNet encodes periodicity by learning a fixed-length recurrent cycle template, both avoiding explicit token-wise temporal interactions. However, due to sharing a single compression module across both spatial units and temporal segments, these methods effectively impose a one-size-fits-all scheme, often homogenizing temporal representations and ignoring heterogeneous temporal patterns across nodes and time periods, risking the loss of temporal diversity and fine-grained dynamics and thus compromising expressiveness and generalization, especially for long-range forecasting tasks.

To address the above limitations, we propose \textbf{\model}, a \underline{Fa}st long-horizon forecasting framework for large-scale \underline{S}patial-\underline{T}emporal graphs. \model is a heterogeneity-aware parallelized  Mixture-of-Experts (MoE) networks. Firstly, an MoE-based temporal compression input module projects historical sequences into a low-dimensional dense embedding, improving computational efficiency and preventing memory explosion concerning the historical input length. To mitigate the information loss and representation homogenization induced by compression, we introduce a heterogeneity-aware router that dynamically selects expert-specific compression pathways across nodes and time periods, thereby extracting diversified temporal features. Furthermore, an expert-concurrent mechanism based on Gated Linear Units (GLUs) is proposed to achieve efficient feature extraction. Secondly, an adaptive graph agent attention module is introduced, where $a$ agent tokens ($a \ll N$) are learned to reduce the long-range spatial interaction complexity from $O(N^2)$ to $O(Na)$. Finally, a multi-layer MoE network is built to extract diverse spatial features. 

The main contributions are summarized as follows:
\begin{itemize}[leftmargin=*]
    \item \textbf{Framework:} We propose \model, the first framework for long-horizon and large-scale STG forecasting that unlocks the prediction span to one week and node size to tens of thousands within a tolerable time.

    \item \textbf{Parallelized Mixture-of-Experts:} A novel heterogeneity-aware router and a parallel GLU-MoE module are proposed, which can efficiently and effectively extract diverse spatial and temporal features based on node heterogeneity.

    \item \textbf{Linear Complexity:} With the Temporal Compression Input with MoE module and the adaptive graph agent attention module, \model achieves linear computational complexity w.r.t. the number of nodes while maintaining the ability of long-range spatial-temporal dependency modeling.

    \item \textbf{Effectiveness:}  Extensive experiments on the large-scale STG forecasting benchmark, as shown in Figure~\ref{fig:intro}, show that \model exhibits superior long-range forecast performance, reducing $4.4\% \sim 18.4\%$ in prediction error, and is $1.3\times \sim 2.2\times$ faster than the SOTA model, demonstrating significant computational efficiency.
\end{itemize}

\section{Problem Formulation} 

\begin{figure*}[ht]
    \centering
    \includegraphics[width=0.95\linewidth]{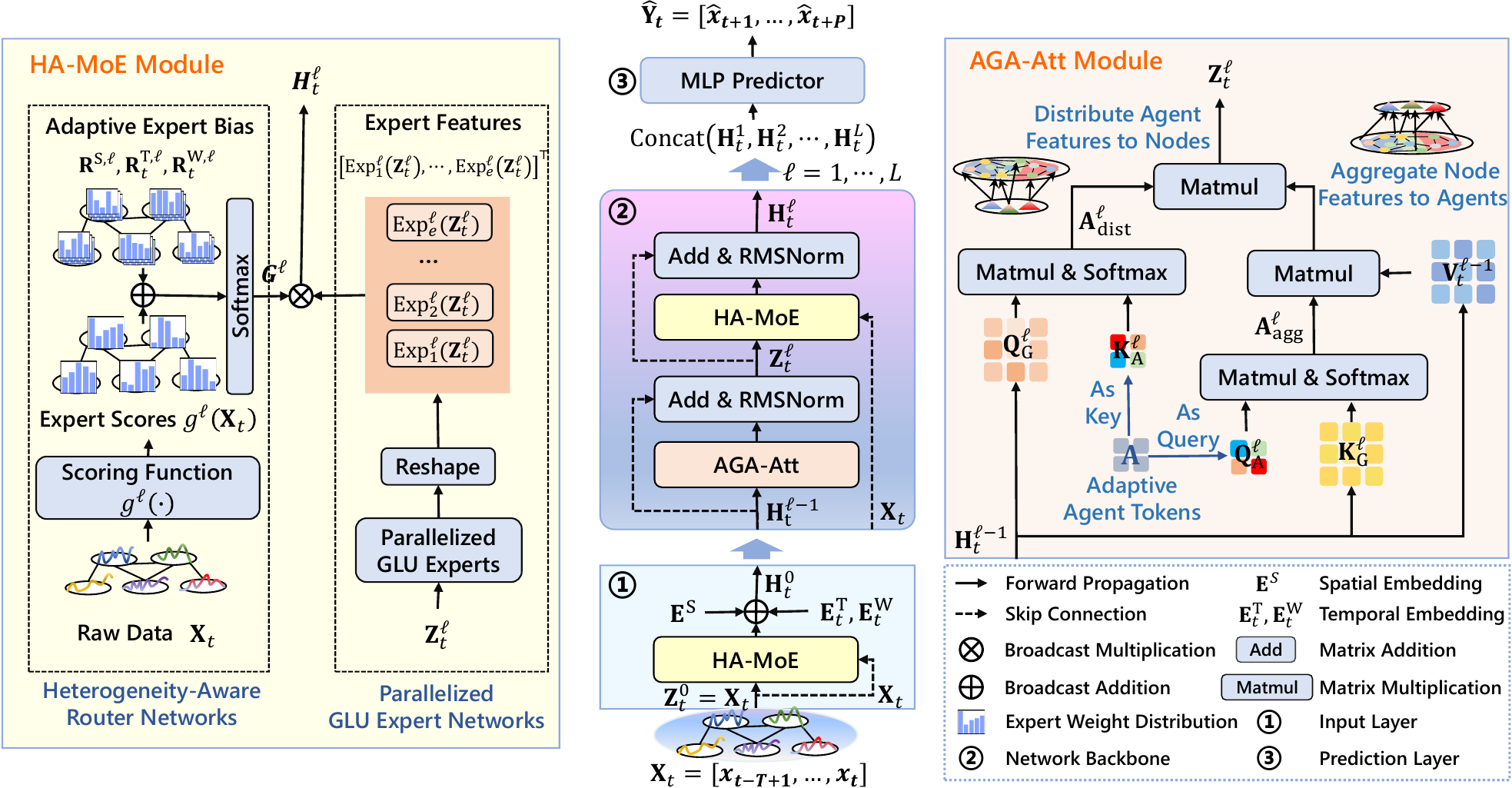}
    \caption{\label{fig:model}
    Architecture of FaST. The middle part illustrates the workflow: the input sequence $\mat{X}_t$ is first embedded and fed into $L$ stacked backbone blocks, and the resulting representations are concatenated and passed to an MLP predictor to generate $\hat{\mat{Y}}_t$.}
    \Description{}
\end{figure*}

Given $N$ spatial nodes (e.g., sensors, regions), we denote measurements (e.g., traffic flow) recorded at \( N \) nodes over \( T \) sequential time steps by \(\mat{X}_t = [\vec{x}_{t-T+1}, \dots, \vec{x}_t] \in \mathbb{R}^{N\times T}\). The relationships among spatial locations are represented using an undirected graph $\mathcal{G}=(\mathcal{V},\mathcal{E})$, where $\mathcal{V}=\{v_i|i\in 1,2,\dots,N\}$ is the set of $N$ nodes, and $\mathcal{E}=\{(v_i,v_j)|v_i,v_j\in \mathcal{V}\}$ is the set of edges among nodes. $\mathcal{E}$ can be derived using a predefined topology structure (e.g., a road network) or can be learned adaptively from the data in an end-to-end manner. In this work, a set $\mathcal{U}=\{u_i|i\in 1,2,\dots,a\}$ of $a$ ($a \ll N$) learnable agent tokens is introduced, whose relationships $\mathcal{E}^\prime=\{(v_i,u_j)|v_i \in \mathcal{V},u_j\in\mathcal{U}\}$ with the original nodes are obtained through interaction. Thus, the graph becomes $\mathcal{G}=(\mathcal{V}\cup\mathcal{U},\mathcal{E}^\prime)$. The goal of a general STG forecasting is to predict the future measurements \(\mat{Y}_t = [\vec{x}_{t+1}, \dots, \vec{x}_{t+P}] \in \mathbb{R}^{N\times P}\) over $P$ time steps by learning a mapping $f_\mathcal{G}$ as follows:
\begin{equation}
\hat{\mat{Y}}_t = f_\mathcal{G}\left(\mat{X}_t;\Theta\right),
\end{equation}
where $\hat{\mat{Y}}$ is the prediction of future traffic flows, and $\Theta$ is the set of all trainable parameters of $f_\mathcal{G}$.

\section{Methodology}
The architecture of \model, as shown in Figure~\ref{fig:model}, consists of three components:
\textcircled{1}~Input Layer: The raw sequence $\mat{X}_t$ is first mapped into dense representations via a Heterogeneity-aware MoE (HA-MoE) module, and then augmented with adaptive spatial and temporal embeddings.
\textcircled{2}~Network Backbone: \model stacks $L$ residual blocks. In each block, a Adaptive Graph Agent Attention (AGA-Att) module first summarizes node features into a small set of agent tokens and redistributes the aggregated information back to all nodes; HA-MoE is then applied to perform heterogeneity-aware enhancement.
\textcircled{3}~Prediction Layer: The outputs from all layers are concatenated and passed to an MLP head to produce the final forecasts.

\subsection{Network Backbone}
\model uses a transformer-like stacked block to perform spatial-temporal feature extraction. Specifically, the proposed AGA-Att is used to replace traditional self-attention, and the proposed \mbox{HA-MoE} is then employed instead of the Feed-Forward Network (FFN) layer to perform efficient feature transformation. The backbone of \model is a stack of $L$ residual blocks, shown as follows:
\begin{equation}
    \begin{aligned}        \mat{H}_t^{\ell}&=\text{RMSNorm}\left(\text{HA-MoE}\left(\mat{Z}_t^{\ell},\mat{X}_t\right) + \mat{Z}_t^{\ell}\right)\\
     \mat{Z}_t^{\ell} &= \text{RMSNorm}\left(\text{AGA-Att}\left(\mat{H}_t^{\ell-1}\right) + \mat{H}_t^{\ell-1}\right),
    \end{aligned}
    \label{eq:backbone}
\end{equation}
where $\text{RMSNorm}(\cdot)$ is a root mean square layer normalization operation~\cite{ZhangS19a}. The input to the first layer ($\ell=1$) of the network is initialized with the features in Eq.~(\ref{eq:input0}).

\subsection{MoE-based Temporal Compression Input}
Long-horizon forecasting requires capturing fine-grained historical dependencies while remaining computationally scalable.
To reconcile accuracy and efficiency, we introduce an MoE-based temporal compression module that projects the $T$-step raw sequence into a compact dense embedding using multiple learnable projection experts.
Unlike one-size-fits-all compression, a heterogeneity-aware router dynamically selects expert-specific compression pathways across nodes and time periods, mitigating representation homogenization and alleviating information loss under compression.

Specifically, given an input $\mat{X}_t \in \mathbb{R}^{N\times T}$, use a HA-MoE module to reduce the information from $T$ time steps to a $d$-dimensional dense embedding as follows:
\begin{equation}
    \mat{\bar H}_t = \text{HA-MoE}\left(\mat{Z}^0_t,\mat{X}_t\right) \in \mathbb{R}^{N\times d},
\end{equation}
where $\mat{Z}^0_t=\mat{X}_t$.
In addition, trainable spatial and temporal embeddings are added to the feature to identify the spatial and temporal positions, shown as follows:
\begin{equation}
    \mat{H}^0_t = \mat{\bar H}_t \oplus \mat{E}^{\mathrm{S}} \oplus \mat{E}_t^{\mathrm{D}} \oplus \mat{E}_t^{\mathrm{W}}\in \mathbb{R}^{N\times d},
    \label{eq:input0}
\end{equation}
where $\oplus$ is a broadcast addition operation, $\mat{E}^{\mathrm{S}}\in \mathbb{R}^{N\times d}$ encodes the structural/positional priors of nodes, and $\mat{E}_t^{\mathrm{D}}\in \mathbb{R}^{1\times d}$ and $\mat{E}_t^{\mathrm{W}}\in \mathbb{R}^{1\times d}$ encode the position of the time step $t$ within a day and a week, respectively. The aforementioned $\mat{E}^{\mathrm{S}}$, $\mat{E}_t^{\mathrm{D}}$, and $\mat{E}_t^{\mathrm{W}}$ are initialized as trainable embeddings and optimized alongside the model.

\subsection{Heterogeneity-Aware MoE (HA-MoE)}
To manifest the impact of heterogeneity on prediction accuracy, it is necessary to characterize this difference through a variety of representations. The mixture of experts architecture is a neural network design that leverages multiple specialized sub-networks (called "\textit{experts}") to handle different parts of the input data ~\cite{MoE}. A gating network dynamically routes each input to the most relevant experts~\cite{ChenDWGL22}, enabling the model to scale efficiently and specialize in diverse tasks ~\cite{ZhouLLDHZDCLL22,Time-MoE}. In \model, the MoE architecture is used to extract diverse features.

\textbf{Load Balancing Problem.} Despite the scalability of MoE architectures, they often suffer from load imbalance, where a small number of experts dominate routing decisions, while others remain underutilized, a phenomenon known as expert weight polarization~\cite{CaiJWTKH25}. Such an imbalance can diminish the model's effective capacity and destabilize training processes. Previous approaches address this issue by implementing auxiliary load balancing losses or capacity constraints~\cite{ShazeerMMDLHD17,Time-MoE}. However, these methods often increase training complexity or restrict the flexibility inherent in expert specialization.  \model tackles this challenge through a Heterogeneity-Aware Router (HA-Router) network and a dense MoE design, enabling all experts to contribute to the forward computation.

\textbf{Heterogeneity-Aware Router Networks (HA-Router).} It avoids expert polarization by exploiting the spatial–temporal heterogeneity of the input. Specifically, HA-Router measures the correlation between experts and features by absorbing raw time-series patterns and injecting adaptive spatial and temporal expert biases. This heterogeneity-aware routing design forces nodes to select experts relevant to themselves rather than concentrating on a single expert. Given $e$ experts and node features $\mat{Z}_t^{\ell}\in \mathbb{R}^{N\times d}$, the expert weight is calculated as follows:
\begin{equation}
    \begin{aligned}
        \mat{G}^{\ell}& = \mathrm{softmax}\left(g^{\ell}(\mat{X}_t)\oplus\mat{R}^{\mathrm{S},\ell}\oplus\mat{R}^{\mathrm{T},\ell}_t\oplus\mat{R}^{\mathrm{W},\ell}_t\right)\in \mathbb{R}^{N\times e}\\
        &g^{\ell}(\mat{X}_t) = \mat{X}_t\mat{W}^{\ell}\in \mathbb{R}^{N\times e},
    \end{aligned}
    \label{eq:Router_scores}
\end{equation}
where $\oplus$ is a broadcast addition operation. $g^{\ell}(\cdot)$ is a scoring function used to calculate expert scores, and $\mat{W}^{\ell}\in \mathbb{R}^{T\times e}$ is a trainable parameter. $\mat{R}^\mathrm{S}\in \mathbb{R}^{N\times e}$ and $\mat{R}^\mathrm{T}_t,\mat{R}^\mathrm{W}_t\in \mathbb{R}^{1\times e}$ are adaptive spatial and temporal expert biases obtained via trainable parameters. Specifically, $\mat{R}^\mathrm{S}$ captures the expert bias on the spatial locations while $\mat{R}^\mathrm{T}_t$ and $\mat{R}^\mathrm{W}_t$ encode the expert bias on the temporal horizons (e.g., time of day and day of week). Finally, the weight distribution $\mat{G}^{\ell}\in \mathbb{R}^{N\times e}$ is used to assign weights to the experts and aggregate their representations, shown as follows:
\begin{equation}
    \begin{aligned}
        \text{HA-MoE}\left(\mat{Z}^{\ell}_t,\mat{X}_t\right) = \sum_{i=1}^{e} \mat{G}^{\ell}[:,i]\otimes \mathrm{Exp}_i(\mat{Z}^{\ell}_t),
    \end{aligned}
    \label{eq:Router}
\end{equation}
where $\otimes$ is a broadcast multiplication, and $\mathrm{Exp}_i(\mat{Z}^{\ell}_t)\in \mathbb{R}^{N\times d}$ is $i$-th expert's output. This router adapts to input variations, making MoE suitable for heterogeneous data. Experts can specialize in different aspects of the data, improving model performance on diverse tasks.

\textbf{Parallelized GLU Expert Networks.} For the selection of experts, existing works generally use FFNs. However, FFNs with a multi-layer structure often require complex mechanisms and high costs to achieve parallelism (e.g., assigning experts to different computing units). A promising alternative is to use linear units as experts, which have good computational efficiency and parallelism. To maintain model capacity and its ability to model nonlinear relationships, GLUs~\cite{GLU} are used as experts in \model. 

Generally, each expert $\mathrm{Exp}_i(\mat{Z}^{\ell}_t)\in \mathbb{R}^{N\times d}$ is an instance of a~GLU:
\begin{equation}
    \begin{aligned}
        \mathrm{Exp}_i(\mat{Z}^{\ell}_t)&=\mathrm{GLU}_i(\mat{Z}^{\ell}_t)\\
        &=\sigma\left(\mat{Z}^{\ell}_t\mat{W}^{\ell}_{i,2} + \vec{b}^{\ell}_{i,2}\right)\odot \left(\mat{Z}^{\ell}_t\mat{W}^{\ell}_{i,1} + \vec{b}^{\ell}_{i,1}\right),
    \end{aligned}
    \label{eq:GLU}
\end{equation}
where $\odot$ denotes the element-wise multiplication, $\sigma$ denotes the sigmoid activation function, $\mat{W}^{\ell}_{i,1}, \mat{W}^{\ell}_{i,2}\in \mathbb{R}^{d\times d}$ and $\vec{b}^{\ell}_{i,1}, \vec{b}^{\ell}_{i,2}\in \mathbb{R}^{d}$ are trainable parameters. However, the instances in Eq.~\ref{eq:GLU} are inefficient, similar to FFNs. Therefore, an efficient parallel computing mechanism for GLU experts is constructed as follows:
\begin{equation}
    \begin{aligned}
        \left[\mathrm{Exp}_1(\mat{Z}^{\ell}_t), \dots, \mathrm{Exp}_e(\mat{Z}^{\ell}_t)\right]^\top&=\mathrm{Reshape}\left(\sigma\left(\mat{F}^{\ell}_1\right)\odot\mat{F}^{\ell}_2\right)\\
        \mat{F}^{\ell}_1,\mat{F}^{\ell}_2&=\mathrm{Split}\left(\mat{Z}^{\ell}_t\mat{W}^{\ell}_{F} + \vec{b}^{\ell}_{F}\right),\\
    \end{aligned}
    \label{eq:ParallGLUMoE}
\end{equation}
where $\top$ indicates a transpose operation. $\mathrm{Reshape}(\cdot)$ and $\mathrm{Split}(\cdot)$ are tensor reshaping and splitting operations, respectively. $\mat{W}^{\ell}_F\in \mathbb{R}^{d\times 2ed}$ and $\vec{b}^{\ell}_F\in \mathbb{R}^{2ed}$ are trainable parameters. As shown in Eq.~(\ref{eq:ParallGLUMoE}), the linear transformation calculations of experts are merged into a linear layer consisting of $\mat{W}_F^{\ell}$ and $\vec{b}_F^{\ell}$. The output of the linear layer is split into two parts ($\mat{F}^{\ell}_1,\mat{F}^{\ell}_2\in \mathbb{R}^{N\times ed}$) to calculate the output of GLUs. Finally, the output $\left[\mathrm{Exp}_1(\mat{Z}^{\ell}_t), \dots, \mathrm{Exp}_e(\mat{Z}^{\ell}_t)\right]^\top\in \mathbb{R}^{N\times e\times d}$ of multiple GLU experts can be calculated simultaneously by reshaping one integrated GLU expert's output from $N\times ed$ to $N\times e\times d$. Therefore, our model can achieve good parallel processing on a single computing unit, avoiding the additional cost of complex parallel mechanisms.

For clarity, the proposed HA-MoE module is denoted as:
\begin{equation}
 \text{HA-MoE}\left(\mat{Z}^{\ell}_t,\mat{X}_t\right) = \mat{G}^{\ell}\otimes\left[\mathrm{Exp}_1(\mat{Z}^{\ell}_t), \dots, \mathrm{Exp}_e(\mat{Z}^{\ell}_t)\right]^\top,
    \label{eq:HA-MoE}
\end{equation}
where $\otimes$ is a broadcast multiplication, and $\mat{G}^{\ell}\in \mathbb{R}^{N\times e}$ is the weight of experts generated by the \text{HA-Router} Network (Eq.~(\ref{eq:Router_scores})).

\subsection{Adaptive Graph Agent Attention (AGA-Att)}
The high complexity of GCNs on raw graph structures prevents a large number of spatial-temporal methods from running on large-scale STGs. Inspired by~\cite{AgentAttention}, \model uses a set of adaptive agent tokens $\mathcal{U}=\{u_i|i\in 1,2,\dots,a\}$ as intermediaries to effectively aggregate and distribute information, capturing long-range spatial dependencies. Specifically, an adaptive embedding token is generated for each agent token, and the embeddings of all agent tokens are denoted as $\mat{A}\in \mathbb{R}^{a\times d}$. Below, the interaction details between agent tokens and graph node tokens are introduced.

\textbf{Node-to-Agent Aggregation Attention:} Each agent token queries all graph node tokens to aggregate long-range spatial information. Concretely, the attention weights from agent queries to graph keys are computed as
\begin{equation}
  \mat{A}^{\ell}_{\mathrm{agg}}=\mathrm{Softmax} \left ( \frac{\mat{Q}^{\ell}_{\mathrm{A}} \mat{K}^{\ell^\mathsf{T}}_{\mathrm{G}}} {\sqrt{d}}\right), \quad \mat{Q}^{\ell}_{\mathrm{A}}=\mat{A}\mat{W}^{\ell,1}_{\mathrm{agg}},\quad \mat{K}^{\ell}_{\mathrm{G}}=\mat{H}^{\ell}_t\mat{W}^{\ell,2}_{\mathrm{agg}},
  \end{equation}
where, $\mat{W}^{\ell,1}_{\mathrm{agg}}$, $\mat{W}^{\ell,2}_{\mathrm{agg}} \in \mathbb{R}^{d\times d}$ are trainable weight matrices.

\textbf{Agent-to-Node Distribution Attention:} Each graph node token queries the agent tokens to obtain the summarized global context and distributes it back to nodes. Specifically, the attention weights from node queries to agent keys are computed as
\begin{equation}
  \mat{A}^{\ell}_{\mathrm{dist}}=\mathrm{Softmax} \left ( \frac{\mat{Q}^{\ell}_{\mathrm{G}} \mat{K}^{\ell^\mathsf{T}}_{\mathrm{A}}} {\sqrt{d}}\right), \quad \mat{Q}^{\ell}_{\mathrm{G}}=\mat{H}^{\ell}_t\mat{W}^{\ell,1}_{\mathrm{dist}},\quad \mat{K}^{\ell}_{\mathrm{A}}=\mat{A}\mat{W}^{\ell,2}_{\mathrm{dist}},
  \end{equation}
where $\mat{W}^{\ell,1}_{\mathrm{dist}}$, $\mat{W}^{\ell,2}_{\mathrm{dist}} \in \mathbb{R}^{d\times d}$ are trainable weight matrices.

The final output of AGA-Att is as follows:
\begin{equation}
    \begin{aligned}
    \text{AGA-Att}(\mat{H}^{\ell-1}_t) &= \mat{A}^{\ell}_{\mathrm{dist}} (\mat{A}^{\ell}_{\mathrm{agg}} \mat{V}^{\ell-1}_t)\\ \mat{V}^{\ell-1}_t&=\mat{H}^{\ell-1}_t\mat{W}^{\ell}_{\mathrm{V}}
    \end{aligned}
\end{equation}
where $\mat{W}^{\ell}_{\mathrm{V}} \in \mathbb{R}^{d\times d}$ is the trainable value projection matrix. The AGA-Att mechanism can be viewed as a low-rank spatial interaction bottleneck: graph tokens first project information to a small set of shared agent tokens and then receive redistributed summaries from them. Apparently, by reducing the number of direct interactions between input tokens, the above mechanism can lower the computational cost from $O(N^2)$ to $O(Na)$, where $N$ is the number of input tokens and $a$ is the number of agent tokens ($a \ll N$). This design is most effective when node-level temporal behaviors contain shared modes (i.e., spatial redundancy), so that the node representation manifold can be well-approximated by a limited number of latent prototypes. When spatial redundancy is weaker, increasing the number of agents can partially compensate, at the expense of higher $O(Na)$ cost.

\subsection{Prediction Layer}\label{subsec:prediction}
In the prediction phase, a multilayer perceptron (MLP) is used to generate future multi-step results directly. Specifically, the features from the $L$ layers are concatenated as follows:
\begin{equation}
    \mat{U}_t = \text{Concat}\left(\mat{H}_t^{1}, \mat{H}_t^{2},\dots,\mat{H}_t^{\ell}\right),
\end{equation}
where $\text{Concat}(\cdot)$~ denotes the concatenate operation. Then $\mat{U}_t \in \mathbb{R}^{N\times Ld}$ is fed into the multilayer perceptron to generate future predictions as follows:
\begin{equation}
\hat{\mat{Y}}_t= \left(\mathrm{ReLU}\left(\mat{U}_t\mat{W}_{1} + \vec{b}_{1}\right)\right)\mat{W}_{2} + \vec{b}_{2},
\label{eq:predict}
\end{equation}
where $\mat{W}_{1}\in \mathbb{R}^{Ld\times Ld}$, $\mat{W}_{2}\in \mathbb{R}^{Ld\times P}$, $\vec{b}_{1}\in \mathbb{R}^{Ld}$, and $\vec{b}_{2}\in \mathbb{R}^{P}$ are trainable parameters.

\subsection{Loss Function}
The Huber loss is chosen as the loss function because it is robust to the regression problem \cite{HuberLoss}.
It is defined as follows:
\begin{equation}
	\mathcal{L}(y, \hat{y})=\left\{\begin{array}{ll}
		\frac{1}{2}(y-\hat{y})^{2} & \text { for }|y-\hat{y}| \leq \delta \\
		\delta|y-\hat{y}|-\frac{1}{2} \delta^{2}, & \text { otherwise,}
	\end{array}\right.
\end{equation}
where $y$ and $\hat{y}$ are the observed and predicted values, and $\delta$ is a threshold (1 by default). The objective function is shown below:
\begin{equation}
	\underset{\Theta}{\arg \min} \sum_{n=1}^{N} \sum_{p=1}^{P}
	\mathcal{L}\left(\mat{Y}_t[n,p], \hat{\mat{Y}}_t[n,p]\right).
	\label{eq:loss}
\end{equation}

\subsection{Complexity Analysis}

The computational complexity of the \model is primarily driven by MoE and attention modules. Since the time dimension is compressed, the space required to store the spatial-temporal hidden representation is $O(Nd)$, and the time complexity of the compression process is $O(Td)$. On this basis, the time and space complexity of calculating and storing $e$ expert representations is $O(ed^2)$ and $O(Ned)$, respectively. For agent attention, the time and space complexity of calculating and storing the attention distribution are $O(Nad)$ and $O(Na)$, respectively. Overall, the temporal complexity of \model is $O(Td+Nad)$ and the spatial complexity is $O(Ned+Na)$, which are linearly related to the number of time steps and nodes, and thus have good scalability. In addition, since the spatial agent tokens avoid pairwise spatial interactions between nodes, \model can achieve good parallelism across multiple computing units.

\begin{figure*}[!ht]
  \centering
  \includegraphics[width = 0.90 \textwidth]{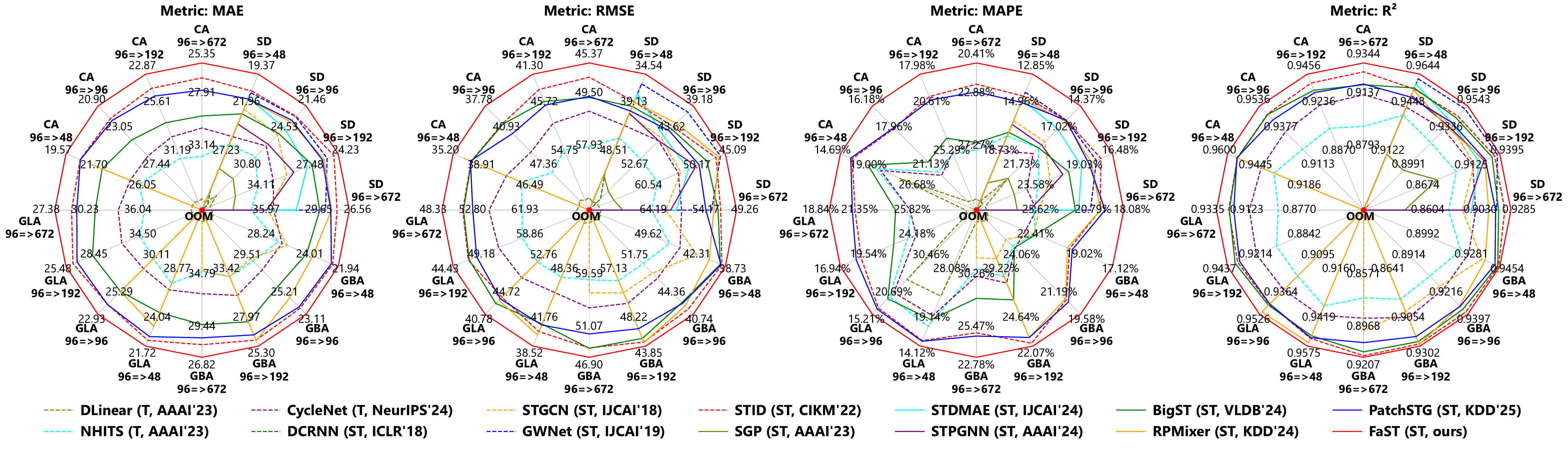}
  \caption{Long-horizon forecasting performance comparison. "96=>672" denotes 672 steps ahead forecasting based on the past 96 time steps. "T" refers to temporal-centric methods, while "ST" denotes spatial-temporal-centric methods. FaST achieved the best performance on 16 prediction tasks.}
  \Description{}
  \label{fig:performance}
\end{figure*}

\section{Experiments}
In this section, we evaluate \model on benchmark datasets. We first introduce the datasets, metrics, and baselines, and then report the main results, ablations, hyperparameter sensitivity, and computational cost. We further provide case studies to show how the MoE design captures heterogeneity across nodes and time periods.

\subsection{Experimental Settings}
\paragraph{Datasets.}
We conduct experiments using the LargeST dataset~\cite{LargeST}, which contains four traffic subset: SD (San Diego), GBA (Greater Bay Area), GLA (Greater Los Angeles), and CA (California). Detailed statistics are provided in Table~\ref{Table:1}. Each subset is split into 6:2:2 training, validation, and test sets in chronological order.

\begin{table}[!htb]
    \caption{\label{Table:1} Dataset statistics.}
    \centering
    \resizebox{0.48\textwidth}{!}{
    \begin{tabular}{c|cccccc}
    \toprule
    \textbf{Data} & \textbf{\#Nodes} & \textbf{Granularity} & \textbf{Time span} & \textbf{Std} & \textbf{Mean} & \textbf{\#Samples}\\
    \midrule
    SD            & 716              & 15 minute            & 2019 (full year) & 184.02       & 244.31        & 24.5M$\sim$25.0M  \\
    GBA           & 2,352            & 15 minute            & 2019 (full year) & 166.67       & 239.82        & 80.6M$\sim$82.1M   \\
    GLA           & 3,834            & 15 minute            & 2019 (full year) & 187.77       & 276.82        & 131.4M$\sim$133.8M \\
    CA            & 8,600            & 15 minute            & 2019 (full year) & 177.12       & 237.39        & 294.7M$\sim$300.1M \\
    \bottomrule
    \end{tabular}
    }
\end{table}

\paragraph{Evaluation Metrics.}
Model performance is evaluated using widely adopted error metrics, including MAE, RMSE, and MAPE, where lower values indicate better predictive accuracy. In addition, we report the coefficient of determination ($R^2$) to quantify how well the predicted sequences explain the variance of the ground-truth series (higher is better).

\paragraph{Baselines.}
To ensure representativeness, timeliness, and competitiveness, we consider two groups of baselines: \textbf{(i) temporal-centric methods} that model time-series dynamics without explicit spatial structures, including
\textit{DLinear}~\cite{DLinear}: decomposes a series into trend and residual components and models them with lightweight linear projections;
\textit{NHITS}~\cite{NHITS}: performs multi-rate processing with hierarchical interpolation for multihorizon forecasting;
and \textit{CycleNet}~\cite{CycleNet}: uses a cycle-consistency objective to better capture periodic patterns.
\textbf{(ii) spatial-temporal-centric methods}, including
\textit{DCRNN}~\cite{DCRNN}: combines diffusion graph convolution with a sequence-to-sequence architecture for spatiotemporal modeling;
\textit{STGCN}~\cite{STGCN}: models spatiotemporal correlations via graph and temporal convolutions;
\textit{GWNet}~\cite{GWNet}: integrates gated temporal convolutions with adaptive graph convolutions;
\textit{SGP}~\cite{SGP}: uses randomized recurrent components with GNNs for scalable learning on large graphs;
\textit{STID}~\cite{STID}: augments simple MLPs with spatial/temporal identity cues to mitigate sample indistinguishability;
\textit{STDMAE}~\cite{STDMAE}: applies decoupled masked autoencoding to reconstruct sequences along spatial and temporal dimensions;
\textit{BigST}~\cite{BigST}: introduces long-sequence feature extraction and linearized global spatial convolution for large-scale networks;
\textit{RPMixer}~\cite{RPMixer}: leverages random projections and a mixer architecture to build a lightweight model;
\textit{STPGNN}~\cite{STPGNN}: selects pivotal nodes and uses parallel convolutions to capture spatiotemporal dependencies in traffic data;
and \textit{PatchSTG}~\cite{PatchSTG}: adopts KDTree-based irregular patching to reduce Transformer cost while preserving fidelity and interpretability.

\paragraph{Implementation Details.}
The task is to forecast the next \{48, 96, 192, 672\} horizons based on observations from the 96 historical time steps (i.e., 1 day). To ensure fairness and reproducibility, all experiments are implemented on the BasicTS~\cite{BasicTS} benchmark framework. All models share the same hidden dimension (\#dim=64), are trained for at most 50 epochs with early stopping, and other baseline hyperparameters follow their official BasicTS configurations. FaST uses a single fixed configuration across all datasets (\#experts=8, \#layers=3, \#agents=32, \#dim=64), selected for efficiency rather than per-dataset tuning. We train FaST with Adam using an initial learning rate of 0.002, decayed by a factor of 0.5 every 10 epochs. All experiments are conducted on an AMD EPYC 7532 @2.40GHz CPU, an NVIDIA RTX A6000 GPU (48GB), and 128GB RAM.

\subsection{Performance Comparison}
Figure \ref{fig:performance} and Table \ref{Table:Performance} (See Appendix) present the long-horizon forecasting results for MAE, RMSE, MAPE, and $R^2$ across specific horizons of 48, 96, 192, and 672.

On the SD dataset, DLinear demonstrates the poorest performance due to its limited capacity for nonlinear modeling. Compared to purely linear baselines (e.g., DLinear), NHITS and CycleNet introduce nonlinear transforms that can better capture non-stationary and multi-scale temporal patterns. Specifically, NHITS employs a hierarchical structure with multiple sampling rates to capture patterns across various time scales. CycleNet, on the other hand, is explicitly designed to model cyclical patterns. These temporal-centric methods cannot effectively leverage spatial correlation, resulting in inferior performance. When modeling spatial-temporal correlations using graphs and convolutions, DCRNN underperforms compared to STGCN, GWNet, and STPGNN. All leverage more advanced GNN architectures, with STGCN excelling in short-range scenarios, GWNet performing better in long-horizon cases, and STPGNN identifying and capturing the spatial-temporal dependencies of pivotal nodes. SGP fails to match its performance due to its less sophisticated approach to capturing spatial dependencies.\par
For large-scale modeling, BigST, STID,  RPMixer, and PatchSTG show improved prediction accuracy, but some methods face scalability challenges on larger datasets. For instance, STGCN and STPGNN cannot complete training within tolerable time frames on GLA/CA, and RPMixer encounters memory limitations on GLA (192-step) and CA (96-step). Consequently, we evaluate only scalable methods, DLinear, NHITS, CycleNet, BigST, STID, PatchSTG, and RPMixer (where feasible), on the larger datasets. The performance trends remain consistent across all datasets: prediction accuracy degrades as horizon length increases. Basically, STID, PatchSTG, and RPMixer outperform BigST on all horizons, confirming their merits in addressing scalability bottlenecks.\par
Regarding \model, it consistently outperforms all baselines across every metric, horizon, and dataset. On SD, it achieves average improvements of 4.73\%, 2.36\%, 8.35\%, and 0.34\% in MAE, RMSE, MAPE, and $R^2$, respectively (vs. best baselines). Significant MAE/MAPE improvements are also observed on GBA (e.g., 1.66\%$\sim$13.40\%), GLA (1.94\%$\sim$8.85\%), and CA (3.67\%$\sim$10.06\%). Despite comparable parameter counts to STID, \model delivers superior performance in all scenarios. Critically, its memory complexity scales linearly with nodes, enabling seamless application to the largest dataset.

\subsection{Ablation Study}
In this section, we conducted ablation experiments by removing different components of the \model model on the four datasets to validate the contribution of each component. For a systematic evaluation, we defined four variants of \model: a) \textbf{w/ LinearInput}: MoE-enhanced input layer is replaced with a linear layer; b) \textbf{w/o HA-MoE}: The HA-MoE module is removed and replaced with GLU; c) \textbf{w/o HA-Router}: The heterogeneity-aware router module is removed, and the expert weight distribution is predicted using the hidden representation output by the previous layer; d) \textbf{w/o AGA-Att}: The adaptive graph agent attention is removed.\par

\begin{figure}[ht]
  \centering
  \includegraphics[width = 0.46\textwidth]{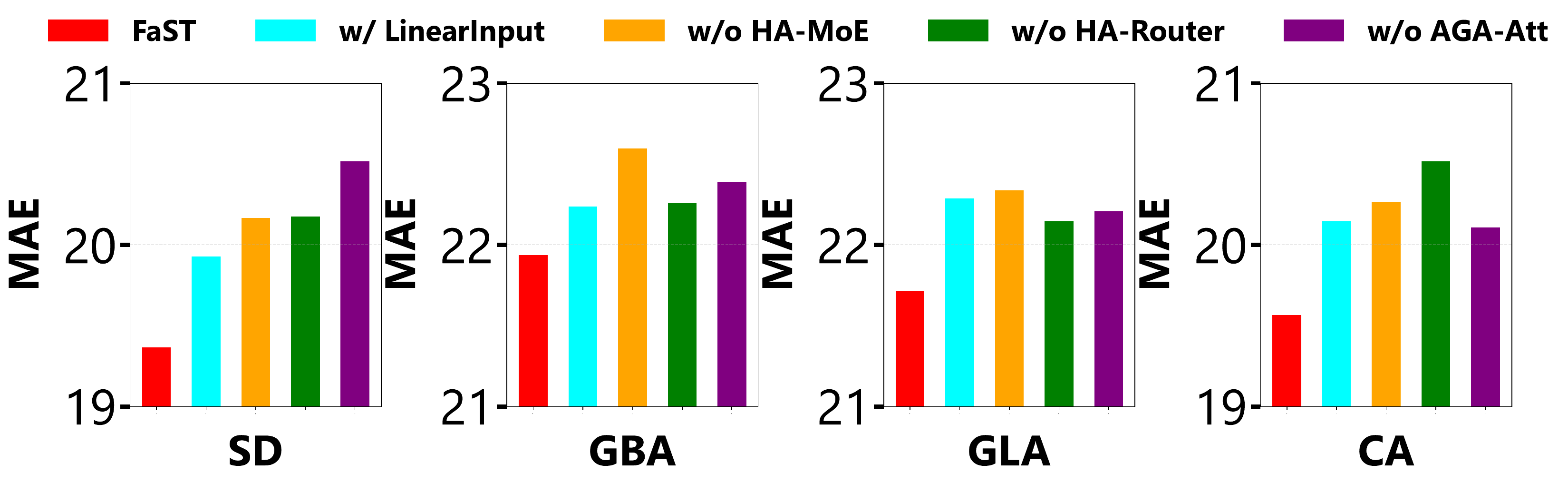}
  \caption{Ablation study of \model components.}
  \Description{}
  \label{fig:ablation}
\end{figure}

As shown in Figure~\ref{fig:ablation}, the complete \model  outperforms all variants on all datasets, validating the effectiveness of each module. Specifically, the \textit{w/ LinearInput} shows a significant performance drop on both the SD and GLA datasets, indicating that the MoE-enhanced input layer is crucial for capturing complex features. The \textit{w/o HA-MoE} exhibits the most pronounced performance decline, especially on GBA, demonstrating the core role of the MoE module in the model. The performance of \textit{w/o HA-Router} and \textit{w/o AGA-Att} also significantly decreases, indicating that the heterogeneity-aware router and agent attention modules contribute importantly to improving model performance. In conclusion, each component of \model plays a positive role in the model's performance, and its complete design achieves optimal performance in complex tasks.

\textbf{Expert Component Comparison.} A standard MoE layer assigns every token to one or more FFNs.  To reduce latency without sacrificing quality, we substitute each FFN expert with a GLU-based expert. Empirically, GLU-Experts match the accuracy of FFN-Experts (Table~\ref{Table:Experts}) while yielding a 1.4x speed-up in wall-clock time.

\begin{table}[ht]
\caption{\label{Table:Experts} \model with GLU-Expert vs. FFN-Expert. Notes: GPU memory usage (G.: GB), number of parameters (P.: K), training time (T.: second/epoch), and inference time (I.: second).}
\centering
\scriptsize
\begin{tabular}{c|c|ccc|cccc}
\toprule
\multirow{2}{*}{Data} & \multirow{2}{*}{Method} & \multicolumn{3}{c|}{Performance} & \multicolumn{4}{c}{Efficiency}  \\
\cmidrule(lr){3-5} \cmidrule(lr){6-9} 
& & MAE & RMSE & MAPE  & \shortstack{G.} & P. & \shortstack{T.} & \shortstack{I.} \\

\midrule
\multirow{2}{*}{SD} 
& FFN-Expert &19.41  & \textbf{34.15} & \textbf{12.80}\% & 2.5& 532 & 15 & 2 \\
& GLU-Expert & \textbf{19.37} & 34.54 & 12.85\% & 1.9 & 549 & 12 & 2\\

\midrule
\multirow{2}{*}{GBA} 
& FFN-Expert & 21.99 & 38.97 & \textbf{17.12}\% & 6.9 & 689 & 48 & 9\\
& GLU-Expert & \textbf{21.94} & \textbf{38.73} & \textbf{17.12}\% & 5.0 & 706 & 36 & 8 \\

\midrule
\multirow{2}{*}{GLA} 
& FFN-Expert & \textbf{21.57} & \textbf{38.26} & \textbf{13.95}\% & 10.7 & 832 & 90 & 15 \\
& GLU-Expert & 21.72 & 38.52 & 14.12\% & 7.6 & 848 & 68 & 14\\

\midrule
\multirow{2}{*}{CA} 
& FFN-Expert & 20.35 & 36.77 & 15.58\% & 23.2 & 1289 & 225 & 37 \\
& GLU-Expert & \textbf{19.57} & \textbf{35.20} & \textbf{14.69}\%  & 17.2 & 1306 & 158 & 32 \\
\bottomrule
\end{tabular}
\end{table}
\subsection{Hyperparameter Sensitivity}
\label{sec:Parameter_Sensitivity}
\begin{figure}[ht]
  \centering
  \includegraphics[width = 0.46\textwidth]{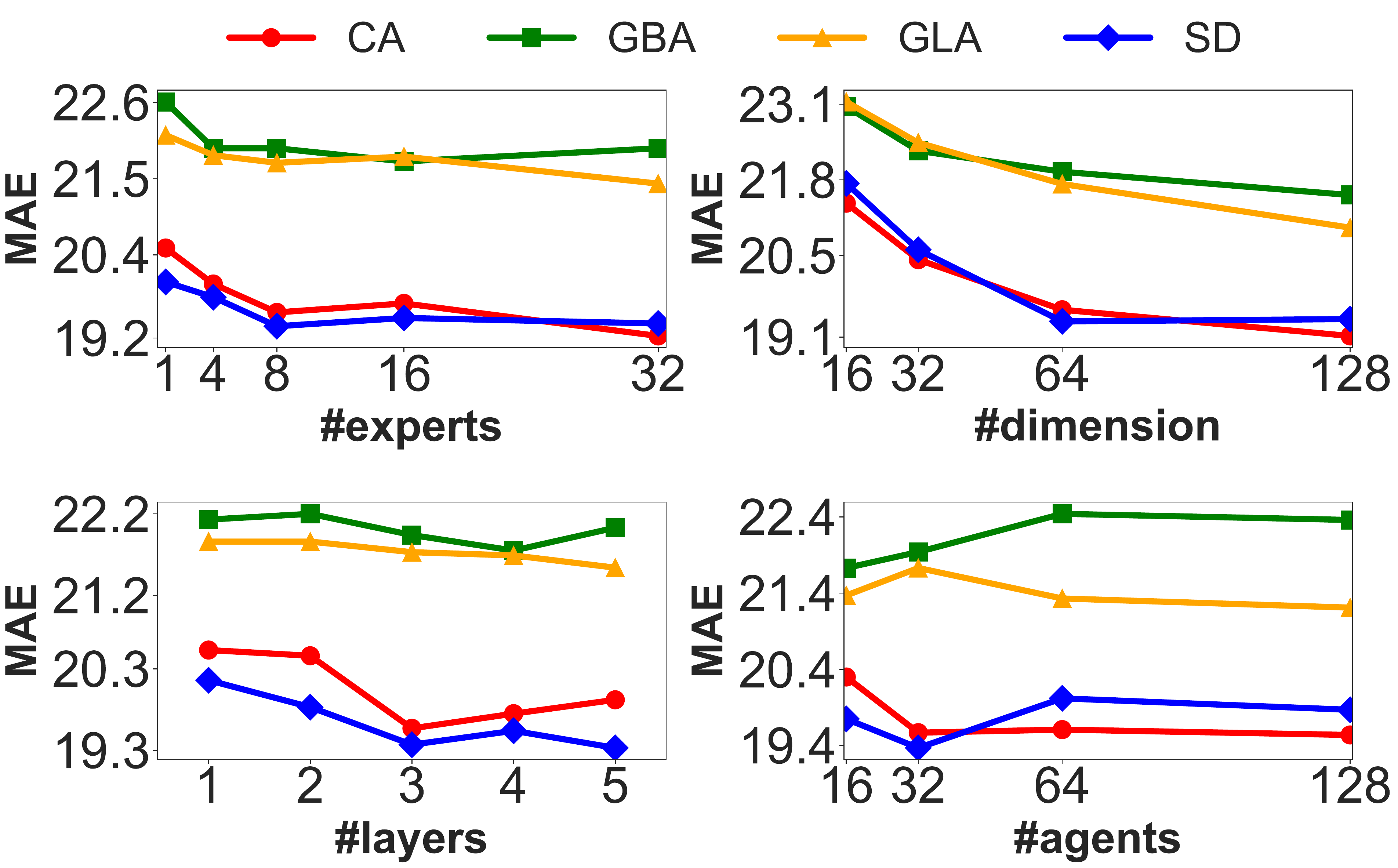}
  \caption{\label{fig:sensitivity_analysis} Visualization of hyperparameter sensitivity.}
  \Description{}
\end{figure}

As illustrated in Figure~\ref{fig:sensitivity_analysis}, we conduct a sensitivity analysis on four critical hyperparameters that significantly influence the model’s performance: the number of experts (\#\textbf{experts}), the number of backbone layers (\#\textbf{layers}), the number of agent tokens in the agent-attention mechanism (\#\textbf{agents}), and the feature dimension (\#\textbf{dimension}). 

\textbf{\#experts}. The optimal number of experts is 8 for  CA/SD and 32 for GBA/GLA. Reducing the number of experts to 4 leads to a notable decline in performance, highlighting the importance of multiple experts in effectively modeling heterogeneity. While increasing the number of experts beyond the optimal values may yield marginal performance improvements, the associated computational costs may outweigh the benefits.

\textbf{\#dimension.} The dimensionality of node embeddings is a crucial hyperparameter that significantly influences model cost, as the majority of the model’s parameters are determined by the number of nodes and the embedding dimension. Increasing the embedding dimension enhances the model’s ability to represent nodes, enabling the capture of more nuanced features. However, this improvement exhibits diminishing returns; for instance, increasing the dimension beyond 64 results in negligible changes in prediction accuracy. 

\textbf{\#layers}. This parameter directly impacts the computational complexity and the number of model parameters. The optimal configuration, however, depends on the dataset size. For the CA and SD datasets, 3 layers achieve the best prediction accuracy, with additional layers leading to overfitting and reduced performance. In contrast, the GBA and GLA datasets benefit from a deeper architecture, with 4 or 5 layers providing a more stable and effective representation learning capacity.

\textbf{\#agents.} We evaluate the agent-attention mechanism with varying numbers of agent tokens: \{16, 32, 64, 128\}. Setting this parameter to 16 or 32 generally delivers robust performance. This range strikes a balance between prediction accuracy and computational efficiency, making it a practical choice for most scenarios.

\subsection{Model Efficiency Comparison}
\begin{figure}[ht]
  \centering
  \subfloat[Model efficiency with varying horizons on CA dataset.]{\includegraphics[width = 0.44\textwidth]{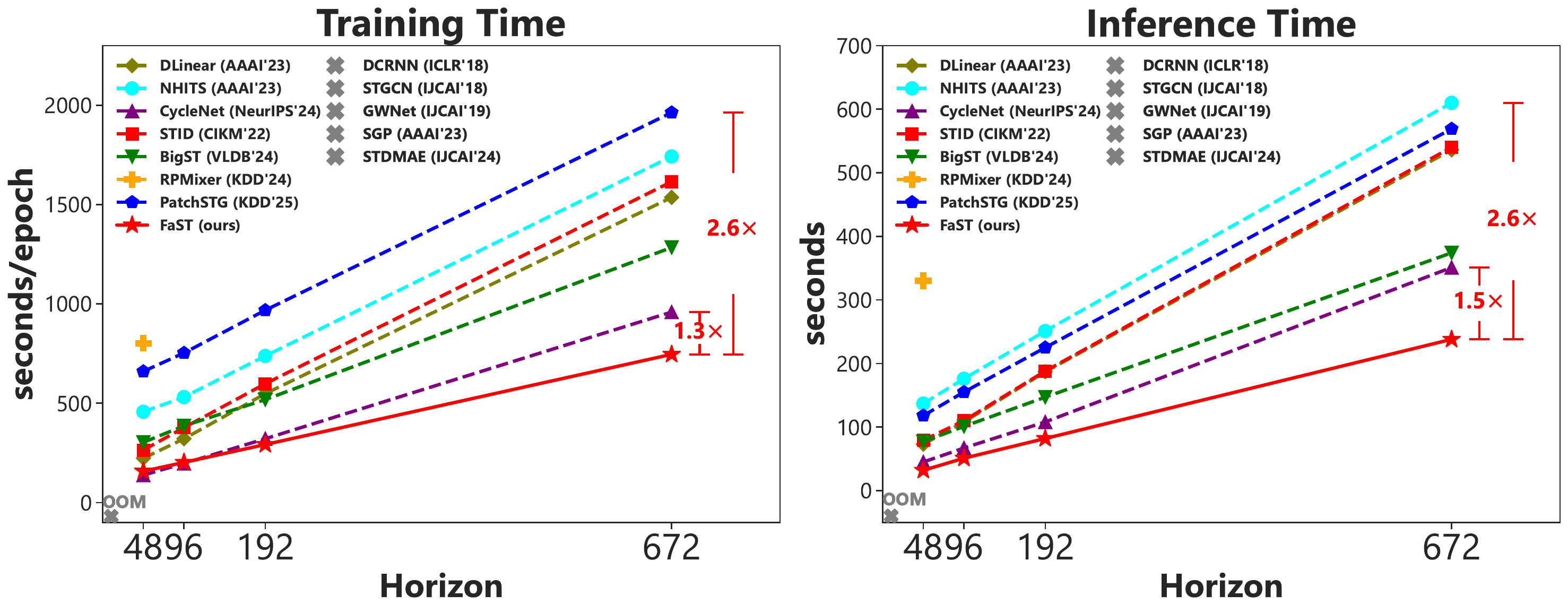}}
  
  \subfloat[Model efficiency with varying graph node scales under 672 horizon.]{\includegraphics[width = 0.44\textwidth]{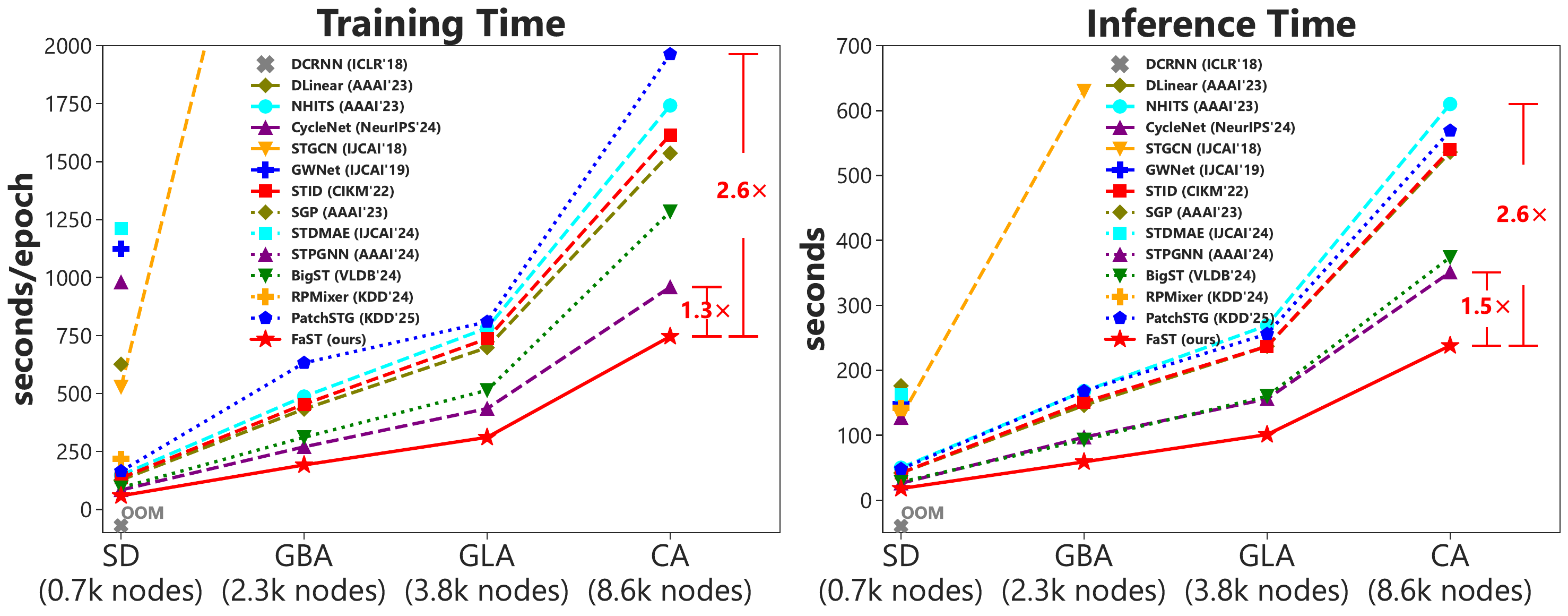}}
   \caption{Model efficiency comparison.}
   \label{fig:efficiency}
\end{figure}

Figure~\ref{fig:efficiency} summarizes the computational efficiency trends across horizons and datasets, while Tables~\ref{Table:Efficiency_horizon} and~\ref{Table:Efficiency_dataset} report the corresponding computational cost in terms of GPU memory usage (G.: GB) and parameter count (P.: K).

\begin{table}[ht]
\caption{\label{Table:Efficiency_horizon}Computational cost across forecasting horizons on the CA dataset. Notes: batch size (B.S.), GPU memory usage (G.: GB), and number of parameters (P.: K).}
\centering
\setlength{\tabcolsep}{2pt}
\resizebox{0.95\linewidth}{!}{
    \begin{tabular}{l|ccc|ccc|ccc|ccc}
    \toprule
    \multirow{2}{*}{\textbf{Method}} &
      \multicolumn{3}{c|}{\textbf{96=>48}} &
      \multicolumn{3}{c|}{\textbf{96=>96}} &
      \multicolumn{3}{c|}{\textbf{96=>192}} &
      \multicolumn{3}{c}{\textbf{96=>672}} \\
    \cmidrule(lr){2-4}\cmidrule(lr){5-7}\cmidrule(lr){8-10}\cmidrule(lr){11-13}
     & B.S. & G. & P. & B.S. & G. & P. & B.S. & G. & P. & B.S. & G. & P. \\
    \midrule
     DLinear & 64 & 3.3 & 9 & 64 & 5.5 & 19 & 64 & 10.3 & 37 & 64 & 32.3 & 130 \\
     NHITS & 64 & 11.6 & 983 & 64 & 13.5 & 1,011 & 64 & 17.2 & 1,066 & 64 & 39.8 & 1,344 \\
     CycleNet & 64 & 5.8 & 900 & 64 & 7.0 & 925 & 64 & 11.6 & 974 & 64 & 33.2 & 1,220 \\
     BigST & 64 & 32.4 & 357 & 64 & 32.8 & 369 & 64 & 38.1 & 394 & 32 & 30.9 & 517 \\
     STID & 64 & 7.9 & 771 & 64 & 10.0 & 778 & 64 & 15.0 & 794 & 64 & 36.7 & 871 \\
     RPMixer & 64 & 27.5 & 7,813 & / & / & / & / & / & / & / & / & / \\
     PatchSTG & 32 & 28.8 & 1,850 & 32 & 29.2 & 1,858 & 32 & 31.4 & 1,874 & 32 & 42.3 & 1,951 \\
     FaST & 64 & 17.2 & 1,306 & 64 & 18.8 & 1,318 & 64 & 21.1 & 1,343 & 64 & 38.1 & 1,466 \\
    \bottomrule
    \end{tabular}
}
\begin{tablenotes}
\footnotesize
\item \dag Methods not included (or marked as '/') encountered GPU out-of-memory errors.
\end{tablenotes}
\end{table}

\begin{table}[ht]
\caption{\label{Table:Efficiency_dataset}
Computational cost across datasets for 672-step forecasting. Notes: batch size (B.S.), GPU memory usage (G.: GB), and number of parameters (P.: K).}
\centering
\setlength{\tabcolsep}{2pt}
\resizebox{0.95\linewidth}{!}{
\begin{tabular}{l|ccc|ccc|ccc|ccc}
\toprule
\multirow{2}{*}{\textbf{Method}} &
  \multicolumn{3}{c|}{\textbf{SD}}  &
  \multicolumn{3}{c|}{\textbf{GBA}} &
  \multicolumn{3}{c|}{\textbf{GLA}} &
  \multicolumn{3}{c}{\textbf{CA}}   \\
\cmidrule(lr){2-4}\cmidrule(lr){5-7}\cmidrule(lr){8-10}\cmidrule(lr){11-13}
  & B.S. & G. & P. & B.S. & G. & P. & B.S. & G. & P. & B.S. & G. & P. \\
\midrule
DLinear   & 64 & 3.2 & 130  & 64 & 9.2 & 130  & 64 & 14.7 & 130  & 64 & 32.3 & 130  \\
NHITS     & 64 & 3.8 & 1,344 & 64 & 11.3 & 1,344 & 64 & 17.9 & 1,344 & 64 & 39.8 & 1,344 \\
CycleNet  & 64 & 3.3 & 463  & 64 & 9.5 & 620  & 64 & 15.1 & 762  & 64 & 33.2 & 1,220 \\
STGCN     & 64 & 38.8 & 1,968 & 32 & 41.1 & 2,806 & / & / & /    & / & / & /    \\
GWNet     & 32 & 34.5 & 888  & / & / & /    & / & / & /    & / & / & /    \\
SGP       & 64 & 6.8 & 999  & / & / & /    & / & / & /    & / & / & /    \\
STDMAE    & 32 & 46.1 & 1,250 & / & / & /    & / & / & /    & / & / & /    \\
BigST     & 64 & 5.7 & 265  & 64 & 17.8 & 317  & 64 & 28.5 & 364  & 32 & 30.9 & 517  \\
STID      & 64 & 3.5 & 366  & 64 & 10.4 & 471  & 64 & 16.6 & 566  & 64 & 36.7 & 871  \\
RPMixer   & 64 & 4.3 & 1,686 & / & / & /    & / & / & /    & / & / & /    \\
STPGNN    & 32 & 32.9 & 762  & / & / & /    & / & / & /    & / & / & /    \\
PatchSTG  & 64 & 8.6 & 2,421 & 64 & 33.9 & 3,291 & 64 & 32.1 & 1,798 & 32 & 42.3 & 1,951 \\
FaST      & 64 & 3.7 & 709  & 64 & 10.9 & 866  & 64 & 17.4 & 1,008 & 64 & 38.1 & 1,466 \\
\bottomrule
\end{tabular}
}
\begin{tablenotes}
\footnotesize
\item \dag Methods not included (or marked as '/') encountered GPU out-of-memory errors.
\end{tablenotes}
\end{table}

For 672-step forecasting on the SD dataset, \model uses only 3.7GB of GPU memory (89.3\% lower than GWNet's 34.5GB) with optimal training (59s/epoch) and inference times (18s). Its parameter count (709K) is higher than DLinear's 130K but less than PatchSTG's 2,421K, balancing performance and efficiency. \model demonstrates strong scalability on the GLA dataset (672-step). Its GPU memory (17.4GB) is 38.9\% lower than BigST's 28.5GB, with training time reduced to 60.5\% of BigST's (514s vs 311s/epoch). The 102.7\% memory growth from 96$\rightarrow$672 step (18.8GB$\rightarrow$38.1GB) is significantly lower than NHITS's 194.8\% (13.5GB$\rightarrow$39.8GB) and DLinear's 487.3\% (5.5GB$\rightarrow$32.3GB), validating its learning efficiency. On the CA dataset (672-step), \model's parameter count (1,466K) exceeds most baselines but achieves competitive training (746s/epoch) and inference times (238s), outperforming BigST in training speed (746s vs 1,284s/epoch) and inference (238s vs 374s). Moreover, the memory growth rate from 96$\rightarrow$672 steps (102.7\%) and training time increase (269\%) are lower than comparable models, confirming strong adaptability to long-sequence tasks.\par

In summary, \model delivers an efficient solution for large-scale time series prediction. By leveraging adaptive agent attention and parallelized expert routing, it achieves high performance with minimal computational overhead. Crucially, its efficiency scales linearly with the number of graph nodes and the length of the horizon, as demonstrated in Figure~\ref{fig:efficiency}.

\subsection{Visualization Analysis} \label{sec:visualization}
\begin{figure}[htbp]
\centering
\subfloat[The distribution of time series.]{\includegraphics[width = 0.4\textwidth]{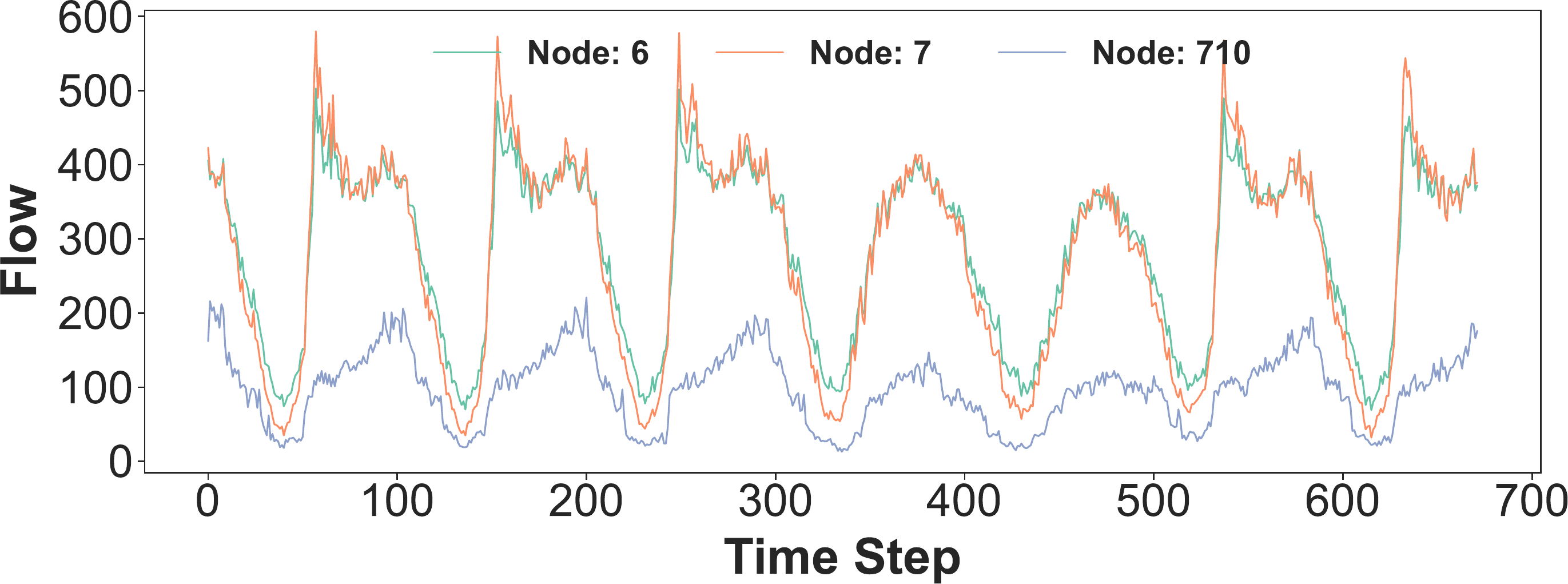}}

\subfloat[Expert weight distribution without HA-Router.]{\includegraphics[width = 0.4\textwidth]{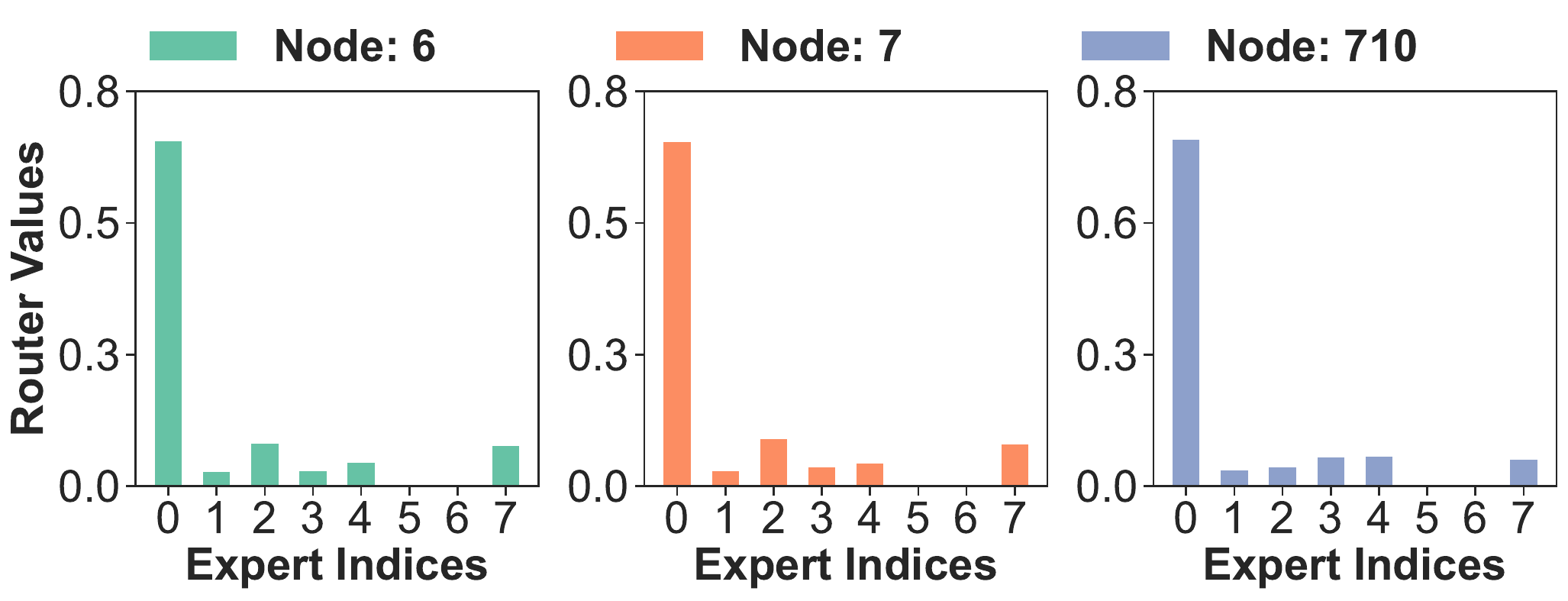}}

\subfloat[Expert weight distribution with HA-Router.]{\includegraphics[width = 0.4\textwidth]{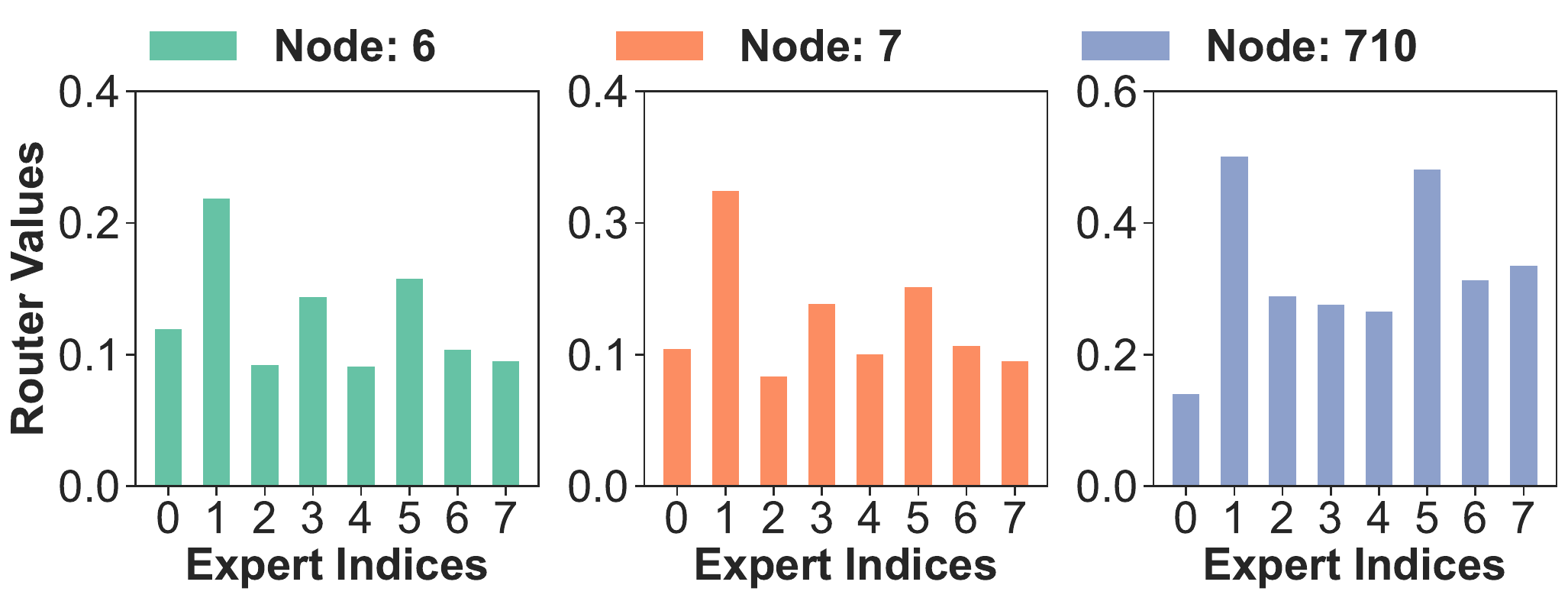}}

\caption{Visualization of expert weight distribution on the SD dataset. The routing distribution generated by HA-Router achieves load balancing.}
\Description{}

\label{fig:vis_expert}
\end{figure}

This section will demonstrate the correlation between traffic patterns and expert weight distribution through case studies.  Using Node 6, Node 7, and Node 710 from the SD dataset as examples, Figure~\ref{fig:vis_expert}a illustrates the cyclical patterns in their time series, while 
Figure~\ref{fig:vis_expert}b displays the distribution of expert weights in the MoE module without HA-Router for these nodes, and  Figure~\ref{fig:vis_expert}c displays the weight distribution in the MoE module for these nodes.

The vanilla MoE module without the HA-Router mechanism suffers from \textit{polarization phenomena}: the majority of weights become randomly concentrated on a single expert. This incurs significant load imbalance and renders routing ineffective. In contrast, the module equipped with the HA-Router mechanism successfully mitigates these adverse effects, avoiding polarization.
Concretely speaking, starting with a horizontal analysis, Node 6 and Node 7, being neighbors, exhibit similar time-series trends, while Node 710 shows a significantly different distribution. Consequently, the expert weights learned for Node 6 and Node 7 are highly consistent: Expert 0 has the largest weight, while Experts 1 and 7 have the smallest and second-smallest weights, respectively. In contrast, the weights for Node 710 are distinctly different. Moving to a longitudinal analysis, slight differences exist between the time-series distributions of Node 6 and Node 7, which are also reflected in the expert weights learned by our model. This shows that the MoE module adapts to the data schema, capturing long-term temporal patterns and spatial heterogeneity among nodes.

\section{Related Work}
\textbf{Spatial-Temporal Graph Forecasting Methods.} Early efforts~\cite{DCRNN, STGCN} utilize GCNs to capture the spatial correlations, and RNNs or CNNs to model temporal dependencies.     Follow-up studies introduce various enhancements, including adaptive graph learning~\cite{GWNet, MTGNN, AGCRN}, dynamic graph learning~\cite{ASTGCN, GMAN, ASTGNN, STPGNN}, spatial-temporal synchronization learning~\cite{STSGCN, STFGNN, STPGCN}, pre-training enhanced methods~\cite{STDMAE}, embedding-based methods~\cite{STID,STAEformer}, and neural ODEs~\cite{STGODE}, to better capture complex spatial and temporal dependencies. These models achieve excellent short-term accuracy (e.g., 12-step) on graphs of a few hundred nodes, but their spatial and temporal modules suffer from quadratic complexity, resulting in excessive memory and time consumption. To address scalability, structure-aware approaches (e.g., SGP~\cite{SGP}, PatchSTG~\cite{PatchSTG}) reduce spatial interactions via sparse aggregation or partitioning, yet often sacrifice long-range dependencies and rely on accurate graph structures. Conversely, structure-free methods (e.g., BigST~\cite{BigST}, RPMixer~\cite{RPMixer}, STID~\cite{STID}) simplify spatial modeling with node embeddings and feature mixing, but may dilute critical spatial semantics. These limitations hinder their effectiveness in large-scale, long-horizon STG forecasting, where preserving rich spatial relationships is essential. To this end, this paper aims to simultaneously improve the accuracy and efficiency of STG in long-term and large-scale prediction.

\textbf{Temporal Forecasting Methods.} STG forecasting can be simplified to a time series forecasting problem by ignoring spatial information. The long-horizon forecasting challenge is tackled by Transformer variants~\cite{Informer,pyraformer,PatchTST,iTransformer}, lightweight MLP-mixers ~\cite{TSMixer,TimeMixer} and linear pattern decomposition~\cite{DLinear,NHITS,CycleNet}. These temporal models lack mechanisms to encode dynamic spatial correlations, limiting their applicability to STG forecasting where joint spatial-temporal modeling is essential.

\textbf{Mixture-of-Experts for Spatial-Temporal Modeling.} MoE models~\cite{MoE} have emerged as a promising paradigm for achieving high model capacity with reduced computation via expert sparsity and input-dependent routing. Although classical MoE architectures were initially developed for static inputs and classification tasks, recent work has extended MoE to time series~\cite{pathformer,Time-MoE,moirai}, where expert specialization and temporal abstraction enable efficient sequence modeling. In spatiotemporal domains, methods like ST-MoE~\cite{ST-MoE} and TESTAM~\cite{testam} apply MoE principles to traffic and weather data, but often rely on sparse gating with limited parallelism. Furthermore, most existing MoE applications overlook the heterogeneity and non-uniform dynamics intrinsic to large-scale urban sensing systems, where different nodes exhibit diverse temporal patterns and interaction semantics. Another challenge is expert load balancing: uneven expert selection leads to polarization, deteriorating both efficiency and accuracy. Addressing this issue is particularly important in dense MoE setups.

\section{Conclusion}
In this paper, we proposed \model, an efficient framework for large-scale long-horizon spatial-temporal forecasting. By integrating a temporal compression input with MoE module, an adaptive graph agent attention mechanism, and a parallelized GLU-MoE module, \model effectively addresses the challenges of forecasting one week ahead with thousands of nodes. The temporal compression input with MoE module extracts heterogeneous features while maintaining scalability, and the adaptive graph agent attention mechanism reduces spatial interaction complexity from quadratic to linear. The parallelized GLU-MoE module further ensures efficient feature extraction with minimal computational overhead. Extensive experiments demonstrate that \model achieves superior long-horizon forecasting performance and significantly improves computational efficiency compared to state-of-the-art models. In future work, we plan to explore the integration of time series foundation models to further enhance the framework’s capabilities and scalability.


\begin{acks}
This work is supported by the National Natural Science Foundation of China (No. 62362069), partially supported by the Yunnan Fundamental Research Projects (No. 202401BF070001-024); the Open Fund of Beijing Key Laboratory of Traffic Data Mining and Embodied Intelligence; and the Open Project Program of Yunnan Key Laboratory of Intelligent Systems and Computing (No. ISC24Y05).
\end{acks}

\newpage

\bibliographystyle{ACM-Reference-Format}
\balance
\bibliography{abrv,ref}

\appendix

\section{Theoretical Analysis of AGA-Att Fidelity}

\subsection{Spatial Reconstruction Error}
AGA-Att can be interpreted as a lossy low-rank projection that compresses node interactions through $a\ll N$ agent tokens.
For a focused analysis of spatial compression, we ignore the value projection $\mat W_{\mathrm{V}}^{\ell}$ (which mainly serves as a learned feature transform) and consider the effective projection:
\begin{equation}
\mat P^{\ell}=\mat A_{\mathrm{dist}}^{\ell}\mat A_{\mathrm{agg}}^{\ell}\in\mathbb{R}^{N\times N}.
\end{equation}
We can define the normalized reconstruction error at layer $\ell$ as
\begin{equation}
\epsilon^{\ell} \;=\; \frac{\left\|\mat H_{t}^{\ell-1}-\mat P^{\ell}\mat H_{t}^{\ell-1}\right\|_{F}}{\left\|\mat H_{t}^{\ell-1}\right\|_{F}},
\label{eq:aga-epsilon}
\end{equation}
which measures how well AGA-Att preserves the input features after spatial compression. Since AGA-Att maps node features to $a$ agent tokens and back, $\mathrm{rank}(\mat P^{\ell}\mat H_{t}^{\ell-1})\le a$ holds, implying a rank-$a$ approximation effect.

\subsection{Fidelity Bounds: Lower and Upper Bounds}

\paragraph{Lower bound.}
Let the singular value decomposition of $\mat H_{t}^{\ell-1}\in\mathbb{R}^{N\times d}$ be $\mat H_{t}^{\ell-1}=\mat U\mat\Sigma\mat V^\top$, with singular values $\sigma_1\ge\cdots\ge\sigma_{\min(N,d)}$.
By the Eckart--Young--Mirsky theorem~\cite{eckart1936approximation}, any rank-$a$ approximation $\hat{\mat H}$ satisfies
$\left\|\mat H_{t}^{\ell-1}-\hat{\mat H}\right\|_F \ge \sqrt{\sum_{i=a+1}^{\min(N,d)}\sigma_i^2}$.
Since $\mat P^{\ell}\mat H_{t}^{\ell-1}$ has rank at most $a$, we obtain
\begin{equation}
\epsilon^{\ell}\;\ge\;\frac{\sqrt{\sum_{i=a+1}^{\min(N,d)}\sigma_i^2}}{\left\|\mat H_{t}^{\ell-1}\right\|_F}.
\label{eq:aga-lb}
\end{equation}
This bound characterizes the \emph{minimum} information loss incurred by any rank-$a$ projection.

\paragraph{Upper bound.}
AGA-Att is closely related to Nystr\"om-style kernel approximation, where agent tokens act as inducing points~\cite{williams2000using,drineas2005nystrom}.
Let $\mat K=\mat H_{t}^{\ell-1}(\mat H_{t}^{\ell-1})^\top$ be the Gram matrix with eigenvalues $\lambda_1\ge\cdots\ge\lambda_N$.
A Nystr\"om approximation typically yields an expected error bounded by the optimal rank-$a$ residual plus a sampling term that decays with $a$. Adapting this to our projection view gives:
\begin{equation}
\mathbb{E}\!\left[\epsilon^{\ell}\right]
\;\le\;
\frac{\sqrt{\sum_{i=a+1}^{N}\lambda_i^2}}{\left\|\mat H_{t}^{\ell-1}\right\|_F}
\;+\;
O\!\left(\frac{1}{\sqrt{a}}\right),
\label{eq:aga-ub}
\end{equation}
suggesting improved fidelity with more agent tokens.

\subsection{Empirical Reconstruction Errors}
\label{app:aga-empirical}
Table~\ref{tab:recon} reports reconstruction errors on the SD dataset under 96$\Rightarrow$48 forecasting while varying the number of agent tokens $a\in\{16,32,64,128\}$.
For the first layer, $\epsilon^{1}$ decreases monotonically as $a$ increases, indicating that early-layer representations contain strong shared spatial modes that can be captured by more agents.
To reflect the cumulative effect across layers, we compute the average reconstruction error $\epsilon_{\mathrm{avg}}$ over all layers, which attains its minimum at $a=32$.
Across the four settings, $\epsilon_{\mathrm{avg}}$ exhibits strong positive correlation with forecasting errors: Pearson correlation coefficients are 0.929 ($p{=}0.071$) with MAE and 0.955 ($p{=}0.045$) with RMSE, implying that higher fidelity (lower $\epsilon_{\mathrm{avg}}$) is associated with better predictive accuracy. Overall, reconstruction errors remain below 0.75, indicating that the approximation is sufficiently faithful for downstream forecasting while enabling scalable computation.

\begin{table}[ht]
\centering
\small
\caption{Reconstruction errors on SD (96$\Rightarrow$48) with different numbers of agent tokens $a$. Lower is better for $\epsilon$.}
\label{tab:recon}
\begin{tabular}{lcccc}
\toprule
\textbf{\#Agent} & $a${=}16 & $a${=}32 & $a${=}64 & $a${=}128 \\
\midrule
$\epsilon^{1}$ ($\ell{=}1$) & 0.611 & 0.512 & 0.488 & 0.484 \\
$\epsilon^{2}$ ($\ell{=}2$) & 0.617 & 0.685 & 0.726 & 0.755 \\
$\epsilon^{3}$ ($\ell{=}3$) & 0.652 & 0.663 & 0.677 & 0.665 \\
$\epsilon_{\text{avg}}$ & 0.627 & \textbf{0.620} & 0.630 & 0.632 \\
\midrule
MAE  & 19.75 & \textbf{19.37} & 20.02 & 19.87 \\
RMSE & 35.22 & \textbf{34.54} & 36.27 & 36.23 \\
\bottomrule
\end{tabular}
\end{table}

\section{Short-Horizon Forecasting Performance}
To evaluate short-horizon forecasting performance, we report results under the 96=>12 setting on SD, GBA, GLA, and CA in Table~\ref{Table:Performance}. FaST achieves the best performance across all datasets. In particular, it improves MAPE by 5.47\% to 11.2\% relative to the strongest baseline, while the relative gains on the other metrics are generally smaller than those observed in long-horizon forecasting.

\section{Generalization to Other Domains}
FaST naturally extends, replacing traffic flows with variables such as electricity demand, precipitation, or temperature. Similar to traffic data, power demand datasets often exhibit high spatial redundancy (e.g., power stations across a region) and temporal heterogeneity (e.g., daily load cycles, seasonal trends). FaST can be applied to power demand forecasting, where the model captures both the spatial correlations between power stations and the temporal dynamics of energy consumption. 

To assess generalization beyond the LargeST benchmarks, we additionally evaluate \model\ on the Electricity (power demand) dataset, which contains 321 variables sampled at 1-hour intervals over 2012--2014. The dataset comprises 8.38M--8.43M samples with mean 2,538.79 and standard deviation 15,027.57. Under the 24-step input to $\{12,24,48,168\}$-step output setting, Table~\ref{tab:electricity} shows that \model consistently achieves the best predictive accuracy, improving MAPE by 12.03\%--19.06\% over the strongest baseline.

\begin{table*}[ht]

\caption{\label{Table:Performance} Performance comparisons. \textbf{Bold} indicates first place, \underline{underline} indicates second place. "96=>48" denotes training on the past 96 time steps to predict the next 48. Missing entries or '/' on GBA, GLA and CA datasets indicate out-of-memory errors.}

\centering

\resizebox{0.98\textwidth}{!}{
\begin{tabular}{c|c|cccc|cccc|cccc|cccc|cccc}
\toprule
\multirow{2}{*}{Data} & \multirow{2}{*}{Method} & \multicolumn{4}{c|}{96=>12} & \multicolumn{4}{c|}{96=>48} & \multicolumn{4}{c|}{96=>96} & \multicolumn{4}{c|}{96=>192} & \multicolumn{4}{c}{96=>672} \\
\cmidrule(lr){3-6} \cmidrule(lr){7-10} \cmidrule(lr){11-14} \cmidrule(lr){15-18} \cmidrule(lr){19-22}
& & MAE & RMSE & MAPE & $R^2$ & MAE & RMSE & MAPE & $R^2$ & MAE & RMSE & MAPE & $R^2$ & MAE & RMSE & MAPE & $R^2$ & MAE & RMSE & MAPE & $R^2$ \\
\midrule

\multirow{15}{*}{SD}
 & DLinear  & 28.32 & 49.90 & 17.57\% & 0.9256 & 36.64 & 65.20 & 23.29\% & 0.8731 & 38.42 & 68.03 & 23.72\% & 0.8623 & 45.95 & 78.12 & 29.04\% & 0.8134 & 46.78 & 80.44 & 31.39\% & 0.8093 \\
 & NHITS    & 19.95 & 33.31 & 12.55\% & 0.9670 & 26.32 & 45.52 & 17.46\% & 0.9369 & 29.13 & 50.84 & 19.44\% & 0.9201 & 34.55 & 58.17 & 23.22\% & 0.8993 & 38.10 & 64.32 & 26.16\% & 0.8780 \\
 & CycleNet & 20.28 & 33.71 & 14.17\% & 0.9653 & 25.40 & 43.07 & 18.66\% & 0.9422 & 27.28 & 46.36 & 20.16\% & 0.9323 & 32.14 & 53.63 & 24.36\% & 0.9144 & 34.80 & 58.26 & 25.92\% & 0.9000 \\
 & DCRNN    & 20.79 & 33.31 & 14.47\% & 0.9668 & 33.23 & 53.54 & 23.22\% & 0.8932 & 42.19 & 65.36 & 27.53\% & 0.8577 & / & / & / & / & / & / & / & / \\
 & STGCN    & 17.69 & 31.81 & 12.38\% & 0.9698 & 22.52 & 41.04 & 16.27\% & 0.9498 & 25.50 & 45.73 & 18.66\% & 0.9377 & 30.27 & 53.16 & 22.11\% & 0.9159 & 35.54 & 61.31 & 25.66\% & 0.8892 \\
 & GWNet    & 16.94 & 28.33 & 11.25\% & 0.9759 & 20.98 & \underline{36.01} & \underline{14.15}\% & \underline{0.9613} & 22.86 & \underline{40.29} & \underline{16.04}\% & \underline{0.9517} & 25.41 & \underline{45.90} & 17.88\% & \underline{0.9373} & 28.86 & 53.07 & 20.94\% & 0.9170 \\
 & SGP      & 20.52 & 35.59 & 12.84\% & 0.9671 & 29.54 & 54.40 & 21.41\% & 0.9107 & 33.71 & 61.54 & 23.51\% & 0.8946 & 38.67 & 70.88 & 25.67\% & 0.8815 & 41.32 & 71.92 & 27.84\% & 0.8425 \\
 & STPGNN   & 17.65 & 29.31 & 11.42\% & 0.9742 & 24.01 & 40.85 & 16.55\% & 0.9502 & 26.98 & 46.35 & 19.07\% & 0.9360 & 29.34 & 50.21 & 20.90\% & 0.9249 & 35.78 & 60.36 & 26.57\% & 0.8926 \\
 & STDMAE   & \underline{16.14} & \underline{27.12} & \underline{10.60}\% & \underline{0.9779} & 21.45 & 37.60 & 14.71\% & 0.9578 & 24.48 & 45.27 & 17.10\% & 0.9390 & 27.70 & 51.64 & 19.19\% & 0.9206 & 32.02 & 59.25 & 21.99\% & 0.8965 \\
 & BigST    & 17.65 & 30.14 & 12.45\% & 0.9740 & 22.98 & 39.67 & 17.16\% & 0.9551 & 25.31 & 43.64 & 19.44\% & 0.9434 & 27.12 & 48.19 & 20.35\% & 0.9324 & 29.45 & 52.06 & 22.28\% & 0.9185 \\
 & STID     & 16.42 & 28.61 & 10.93\% & 0.9746 & \underline{20.82} & 38.87 & 14.44\% & 0.9545 & \underline{22.82} & 43.23 & \underline{16.04}\% & 0.9450 & \underline{24.68} & 46.66 & \underline{17.48}\% & 0.9348 & \underline{27.72} & \underline{51.65} & 20.09\% & \underline{0.9235} \\
 & RPMixer  & 17.11 & 28.51 & 11.19\% & 0.9743 & 21.60 & 38.90 & 14.66\% & 0.9538 & 23.59 & 42.54 & 16.17\% & 0.9348 & 25.72 & 46.22 & 18.31\% & 0.9255 & 28.26 & 52.74 & 20.08\% & 0.9105 \\
 & PatchSTG & 16.29 & 27.67 & 12.47\% & 0.9771 & 21.55 & 40.45 & 14.89\% & 0.9490 & 23.38 & 44.07 & 16.06\% & 0.9392 & 25.91 & 49.64 & 18.54\% & 0.9262 & 28.33 & 54.02 & \underline{19.67}\% & 0.9161 \\
 & \model   & \textbf{15.77} & \textbf{27.09} & \textbf{10.02}\% & \textbf{0.9781}
            & \textbf{19.37} & \textbf{34.54} & \textbf{12.85}\% & \textbf{0.9644}
            & \textbf{21.46} & \textbf{39.18} & \textbf{14.37}\% & \textbf{0.9543}
            & \textbf{24.23} & \textbf{45.09} & \textbf{16.48}\% & \textbf{0.9395}
            & \textbf{26.56} & \textbf{49.26} & \textbf{18.08}\% & \textbf{0.9285} \\
\cmidrule{2-22}
 & \textit{Improvement} & 2.29\% & 0.11\% & 5.47\% & 0.02\% & 6.96\% & 4.08\% & 9.19\% & 0.32\% & 5.96\% & 2.76\% & 10.41\% & 0.27\% & 1.82\% & 1.76\% & 5.72\% & 0.23\% & 4.18\% & 4.63\% & 8.08\% & 0.54\% \\
\midrule

\multirow{10}{*}{GBA}
 & DLinear  & 28.21 & 47.95 & 22.20\% & 0.9161 & 34.95 & 61.09 & 26.25\% & 0.8646 & 36.20 & 63.16 & 27.08\% & 0.8552 & 42.44 & 71.58 & 34.47\% & 0.8145 & 43.38 & 72.73 & 35.77\% & 0.8094 \\
 & NHITS    & 20.74 & 34.99 & 15.72\% & 0.9552 & 26.53 & 46.04 & 22.39\% & 0.9222 & 28.68 & 49.75 & 24.79\% & 0.9088 & 32.38 & 54.60 & 28.38\% & 0.8903 & 34.98 & 58.36 & 30.72\% & 0.8758 \\
 & CycleNet & 21.36 & 35.34 & 17.84\% & 0.9538 & 26.19 & 44.51 & 23.10\% & 0.9270 & 27.57 & 46.82 & 24.57\% & 0.9197 & 30.19 & 50.47 & 27.80\% & 0.9060 & 32.50 & 53.70 & 29.97\% & 0.8911 \\
 & STGCN    & 20.45 & 34.09 & 16.30\% & 0.9575 & 25.78 & 42.24 & 22.47\% & 0.9350 & 28.96 & 47.81 & 25.34\% & 0.9170 & 32.38 & 52.33 & 30.21\% & 0.9006 & 35.07 & 56.09 & 31.79\% & 0.9006 \\
 & BigST    & 19.67 & 32.32 & 16.12\% & 0.9629 & 24.18 & 39.32 & 22.34\% & 0.9442 & 25.80 & 42.56 & 24.63\% & 0.9357 & 27.51 & 44.96 & 26.20\% & 0.9264 & 29.54 & \underline{48.03} & 27.91\% & 0.9166 \\
 & STID     & 18.13 & 31.46 & 14.80\% & \underline{0.9645} & \underline{22.31} & \underline{39.14} & 19.95\% & \underline{0.9447} & \underline{23.65} & \underline{41.54} & 20.61\% & \underline{0.9380} & 25.82 & 44.43 & \underline{22.32}\% & 0.9284 & \underline{27.83} & 48.09 & 24.86\% & \underline{0.9195} \\
 & RPMixer  & 18.68 & 31.11 & 14.69\% & 0.9623 & 23.73 & 40.65 & 20.09\% & 0.9368 & 24.64 & 42.03 & 20.68\% & 0.9330 & \underline{25.77} & \underline{44.35} & 22.81\% & \underline{0.9288} & / & / & / & / \\
 & PatchSTG & \underline{18.12} & \textbf{30.59} & \underline{13.93}\% & 0.9641 & 22.40 & 39.23 & \underline{19.77}\% & 0.9426 & 23.76 & 42.45 & \underline{20.45}\% & 0.9334 & 26.29 & 46.39 & 22.82\% & 0.9221 & 28.40 & 49.94 & \underline{24.58}\% & 0.9096 \\
 & \model   & \textbf{17.73} & \underline{30.71} & \textbf{12.37}\% & \textbf{0.9656}
            & \textbf{21.94} & \textbf{38.73} & \textbf{17.12}\% & \textbf{0.9454}
            & \textbf{23.11} & \textbf{40.74} & \textbf{19.58}\% & \textbf{0.9397}
            & \textbf{25.30} & \textbf{43.85} & \textbf{22.07}\% & \textbf{0.9302}
            & \textbf{26.82} & \textbf{46.90} & \textbf{22.78}\% & \textbf{0.9207} \\
\cmidrule{2-22}
 & \textit{Improvement} & 2.15\% & -0.39\% & 11.20\% & 0.11\% & 1.66\% & 1.05\% & 13.40\% & 0.07\% & 2.28\% & 1.93\% & 4.25\% & 0.18\% & 1.82\% & 1.13\% & 1.12\% & 0.15\% & 3.63\% & 2.35\% & 7.32\% & 0.13\% \\
\midrule

\multirow{9}{*}{GLA}
 & DLinear  & 29.04 & 49.83 & 17.80\% & 0.9295 & 36.63 & 58.28 & 22.88\% & 0.8849 & 38.03 & 65.63 & 23.83\% & 0.8772 & 44.88 & 74.97 & 29.74\% & 0.8396 & 45.61 & 76.27 & 31.07\% & 0.8346 \\
 & NHITS    & 20.77 & 34.82 & 12.39\% & 0.9653 & 27.15 & 47.00 & 17.47\% & 0.9360 & 29.51 & 51.00 & 19.44\% & 0.9250 & 33.95 & 56.89 & 23.12\% & 0.9063 & 36.77 & 61.14 & 25.57\% & 0.8919 \\
 & CycleNet & 21.17 & 35.21 & 14.22\% & 0.9638 & 26.50 & 44.81 & 19.34\% & 0.9411 & 28.13 & 47.51 & 20.52\% & 0.9350 & 31.55 & 52.11 & 23.96\% & 0.9216 & 33.54 & 55.23 & 25.54\% & 0.9113 \\
 & BigST    & 19.05 & 31.21 & 13.24\% & 0.9698 & 24.41 & 40.94 & 18.67\% & 0.9520 & 25.33 & \underline{42.62} & 18.98\% & 0.9461 & 28.17 & 46.97 & 21.79\% & \underline{0.9379} & 31.10 & 51.32 & 24.26\% & 0.9248 \\
 & STID     & 17.53 & 29.97 & 11.19\% & \underline{0.9741} & \underline{22.15} & 40.08 & 15.05\% & 0.9534 & 23.93 & 43.71 & 16.69\% & 0.9454 & \underline{26.16} & \underline{46.84} & 18.78\% & 0.9360 & \underline{28.85} & \underline{51.22} & 20.89\% & \underline{0.9252} \\
 & RPMixer  & 18.48 & 30.15 & 11.50\% & 0.9720 & 23.26 & \underline{39.71} & 15.78\% & \underline{0.9557} & 25.34 & 43.17 & 19.15\% & \underline{0.9509} & / & / & / & / & / & / & / & / \\
 & PatchSTG & \underline{17.51} & \underline{29.79} & \underline{11.03}\% & \underline{0.9741} & 22.44 & 40.83 & \underline{14.94}\% & 0.9530 & \underline{23.91} & 43.42 & \underline{16.42}\% & 0.9452 & 26.57 & 48.20 & \underline{18.34}\% & 0.9345 & 29.33 & 52.69 & \underline{20.67}\% & 0.9218 \\
 & \model   & \textbf{17.20} & \textbf{29.22} & \textbf{10.27}\% & \textbf{0.9756}
            & \textbf{21.72} & \textbf{38.52} & \textbf{14.12}\% & \textbf{0.9575}
            & \textbf{22.93} & \textbf{40.78} & \textbf{15.21}\% & \textbf{0.9526}
            & \textbf{25.48} & \textbf{44.43} & \textbf{16.94}\% & \textbf{0.9437}
            & \textbf{27.38} & \textbf{48.33} & \textbf{18.84}\% & \textbf{0.9335} \\
\cmidrule{2-22}
 & \textit{Improvement} & 1.77\% & 1.91\% & 6.89\% & 0.15\% & 1.94\% & 3.00\% & 5.49\% & 0.19\% & 4.10\% & 4.32\% & 7.37\% & 0.18\% & 2.60\% & 5.15\% & 7.63\% & 0.62\% & 5.10\% & 5.64\% & 8.85\% & 0.90\% \\
\midrule

\multirow{9}{*}{CA}
 & DLinear  & 26.51 & 46.11 & 19.10\% & 0.9314 & 33.34 & 59.03 & 23.66\% & 0.8876 & 34.63 & 56.98 & 24.72\% & 0.8796 & 40.83 & 69.78 & 30.87\% & 0.8431 & 41.65 & 71.09 & 32.34\% & 0.8380 \\
 & NHITS    & 19.57 & 33.21 & 13.90\% & 0.9644 & 25.61 & 44.80 & 20.04\% & 0.9353 & 27.99 & 48.91 & 21.72\% & 0.9229 & 32.10 & 54.32 & 25.49\% & 0.9049 & 34.28 & 57.83 & 27.52\% & 0.8928 \\
 & CycleNet & 19.96 & 33.58 & 15.64\% & 0.9636 & 24.69 & 42.37 & 20.69\% & 0.9421 & 26.18 & 44.78 & 22.03\% & 0.9353 & 29.27 & 49.10 & 25.34\% & 0.9223 & 31.05 & 51.93 & 26.85\% & 0.9136 \\
 & BigST    & 17.65 & 29.32 & 14.07\% & 0.9723 & 22.26 & 37.47 & 18.76\% & 0.9547 & 24.28 & 40.29 & 20.81\% & 0.9477 & 27.64 & 45.44 & 23.92\% & 0.9335 & 29.86 & 49.93 & 26.88\% & 0.9201 \\
 & STID     & 16.32 & 28.27 & 12.46\% & 0.9742 & \underline{20.58} & 37.43 & 16.42\% & 0.9548 & \underline{22.07} & \underline{40.22} & 17.95\% & \underline{0.9479} & \underline{24.16} & \underline{43.55} & \underline{19.99}\% & \underline{0.9389} & \underline{26.52} & \underline{47.18} & \underline{22.06}\% & \underline{0.9286} \\
 & RPMixer  & 17.33 & 28.51 & 13.37\% & 0.9719 & 21.48 & \underline{36.54} & 17.09\% & \underline{0.9563} & / & / & / & / & / & / & / & / & / & / & / & / \\
 & PatchSTG & \underline{16.15} & \underline{27.80} & \underline{12.21}\% & \underline{0.9753} & 20.59 & 37.34 & \underline{16.21}\% & 0.9550 & 22.25 & 41.72 & \underline{17.90}\% & 0.9439 & 24.89 & 46.00 & 20.30\% & 0.9318 & 27.54 & 49.77 & 22.67\% & 0.9206 \\
 & \model   & \textbf{15.78} & \textbf{27.37} & \textbf{10.95}\% & \textbf{0.9758}
            & \textbf{19.57} & \textbf{35.20} & \textbf{14.69}\% & \textbf{0.9600}
            & \textbf{20.90} & \textbf{37.78} & \textbf{16.18}\% & \textbf{0.9536}
            & \textbf{22.87} & \textbf{41.30} & \textbf{17.98}\% & \textbf{0.9456}
            & \textbf{25.35} & \textbf{45.37} & \textbf{20.41}\% & \textbf{0.9344} \\
\cmidrule{2-22}
 & \textit{Improvement} & 2.29\% & 1.55\% & 10.32\% & 0.05\% & 4.91\% & 3.67\% & 9.38\% & 0.39\% & 5.30\% & 6.07\% & 9.61\% & 0.60\% & 5.34\% & 5.17\% & 10.06\% & 0.71\% & 4.41\% & 3.84\% & 7.48\% & 0.62\% \\
\bottomrule
\end{tabular}
}
\end{table*}

\begin{table*}[ht]
\caption{\label{tab:electricity} Performance comparisons on the Electricity dataset (24=>12/24/48/168). \textbf{Bold} indicates first place, \underline{underline} indicates second place. The notation "24=>12" denotes training on the past 24 time steps to predict the next 12 time steps.}
\centering
\resizebox{0.98\textwidth}{!}{
\begin{tabular}{c|c|cccccccccccc|c}
\toprule
Task& Metric & DLinear & NHITS & CycleNet & DCRNN & STGCN & GWNet & SGP & STPGNN & STDMAE & BigST & STID & FaST & \textit{Improv.} \\
\midrule
\multirow{4}{*}{\makecell[c]{Electricity\\24=>12}} & MAE & 195.86 & 223.07 & 181.52 & 218.80 & 236.69 & 212.68 & 508.48 & 234.22 & 200.33 & 197.18 & \underline{165.82} & \textbf{160.61} & 3.14\% \\
 & RMSE & 1663.77 & 1673.13 & 1512.69 & 1926.86 & 2155.44 & 2036.92 & 5925.27 & 1908.38 & 1809.45 & 1606.10 & \underline{1443.72} & \textbf{1420.11} & 1.64\% \\
 & MAPE & 13.26\% & 15.86\% & 12.17\% & 24.90\% & 17.12\% & 16.19\% & 46.71\% & 24.51\% & 15.24\% & 16.33\% & \underline{11.78\%} & \textbf{9.55\%} & 18.93\% \\
 & $R^2$ & 0.9901 & 0.9900 & 0.9918 & 0.9867 & 0.9834 & 0.9852 & 0.8730 & 0.9870 & 0.9883 & 0.9908 & \underline{0.9925} & \textbf{0.9928} & 0.03\% \\
\midrule
\multirow{4}{*}{\makecell[c]{Electricity\\24=>24}} & MAE & 198.88 & 245.26 & 189.61 & 230.44 & 283.29 & 221.44 & 550.38 & 252.49 & 206.87 & 208.84 & \underline{176.46} & \textbf{170.47} & 3.39\% \\
 & RMSE & 1692.41 & 1787.36 & 1597.95 & 1901.23 & 2652.54 & 2102.89 & 7056.51 & 1968.79 & 1815.60 & 1649.40 & \underline{1547.25} & \textbf{1519.00} & 1.83\% \\
 & MAPE & 13.42\% & 18.50\% & 12.70\% & 29.78\% & 20.69\% & 16.04\% & 43.46\% & 23.25\% & 15.27\% & 20.94\% & \underline{12.99\%} & \textbf{10.28\%} & 20.86\% \\
 & $R^2$ & 0.9898 & 0.9886 & 0.9909 & 0.9871 & 0.9749 & 0.9842 & 0.8200 & 0.9862 & 0.9882 & 0.9903 & \underline{0.9914} & \textbf{0.9918} & 0.04\% \\
\midrule
\multirow{4}{*}{\makecell[c]{Electricity\\24=>48}} & MAE & 225.99 & 266.36 & 216.46 & 241.67 & 296.59 & 232.28 & 518.43 & 260.20 & 233.52 & 232.00 & \underline{198.44} & \textbf{195.33} & 1.57\% \\
 & RMSE & 1967.40 & 2011.66 & 1877.68 & 2089.45 & 2512.15 & 2106.64 & 5674.09 & 2062.40 & 2058.08 & 1929.12 & \underline{1817.27} & \textbf{1773.55} & 2.41\% \\
 & MAPE & 15.69\% & 20.65\% & 14.88\% & 18.64\% & 20.11\% & 17.17\% & 59.00\% & 23.33\% & 16.99\% & 21.75\% & \underline{13.64\%} & \textbf{11.96\%} & 12.32\% \\
 & $R^2$ & 0.9862 & 0.9856 & 0.9874 & 0.9821 & 0.9775 & 0.9842 & 0.8837 & 0.9848 & 0.9849 & 0.9867 & \underline{0.9882} & \textbf{0.9888} & 0.06\% \\
\midrule
\multirow{4}{*}{\makecell[c]{Electricity\\24=>168}} & MAE & 261.17 & 303.28 & 253.98 & 268.28 & 309.52 & 256.86 & 580.57 & 292.39 & 261.30 & 268.86 & \underline{239.54} & \textbf{238.77} & 0.32\% \\
 & RMSE & 2633.81 & 2661.56 & 2567.02 & 2684.09 & 3089.76 & 2489.71 & 7123.25 & 2709.39 & 2553.83 & 2581.57 & \underline{2487.66} & \textbf{2446.63} & 1.65\% \\
 & MAPE & 16.81\% & 22.01\% & 16.30\% & 17.08\% & 21.01\% & 16.64\% & 49.29\% & 21.89\% & 19.07\% & 22.34\% & \underline{15.88\%} & \textbf{13.97\%} & 12.03\% \\
 & $R^2$ & 0.9753 & 0.9748 & 0.9766 & 0.9745 & 0.9661 & 0.9762 & 0.8175 & 0.9739 & 0.9768 & 0.9763 & \underline{0.9780} & \textbf{0.9787} & 0.07\% \\
\bottomrule
\end{tabular}%
}
\begin{tablenotes}
\footnotesize
\item \dag RPMixer and PatchSTG are not reported because this dataset does not provide latitude/longitude information required by these methods for spatial partitioning/patch construction.
\end{tablenotes}

\end{table*}

\end{document}